\documentclass[review,12pt,3p]{elsarticle}

\usepackage{lineno}
\usepackage[flushleft]{threeparttable}
\usepackage{fixltx2e}
\usepackage{amsmath, amssymb}

\newcommand{\norm}[1]{\left\lVert#1\right\rVert}

\newcommand{\ra}[1]{\renewcommand{\arraystretch}{#1}}

\modulolinenumbers[5]
\usepackage{setspace}
\usepackage{graphicx}
\usepackage{verbatim} 
\usepackage{amsmath}
\usepackage{times} 
\usepackage{multirow}
\usepackage{amsfonts}
\usepackage[T1]{fontenc}
\usepackage{url}
\usepackage{amssymb}
\usepackage[caption=false,font=footnotesize]{subfig}
\usepackage{epsfig}
\renewcommand{\arraystretch}{1.385}
\usepackage[framemethod=tikz]{mdframed}

\definecolor{cccolor}{rgb}{.67,.7,.67}

\journal{Pattern Recognition}
\begin{document}

\begin{frontmatter}

\title{Learning Features for Offline Handwritten Signature Verification using Deep Convolutional Neural Networks}		

\author[ets]{Luiz G. Hafemann\corref{mycorrespondingauthor}}
\ead{lghafemann@livia.etsmtl.ca}
\author[ets]{Robert Sabourin}
\ead{robert.sabourin@etsmtl.ca}

\author[ufpr]{Luiz S. Oliveira}
\ead{lesoliveira@inf.ufpr.br}

\address[ets]{LIVIA, \'{E}cole de Technologie Sup\'{e}rieure, University of Quebec, Montreal, Quebec, Canada}

\address[ufpr]{Department of Informatics, Federal University of Parana (UFPR), Curitiba, PR, Brazil}

\cortext[mycorrespondingauthor]{Corresponding author}

\begin{abstract}	

Verifying the identity of a person using handwritten signatures is challenging in the presence of skilled forgeries, where a forger has access to a person's signature and deliberately attempt to imitate it. In offline (static) signature verification, the dynamic information of the signature writing process is lost, and it is difficult to design good feature extractors that can distinguish genuine signatures and skilled forgeries. This reflects in a relatively poor performance, with verification errors around 7\% in the best systems in the literature. To address both the difficulty of obtaining good features, as well as improve system performance, we propose learning the representations from signature images, in a Writer-Independent format, using Convolutional Neural Networks. In particular, we propose a novel formulation of the problem that includes knowledge of skilled forgeries from a subset of users in the feature learning process, that aims to capture visual cues that distinguish genuine signatures and forgeries regardless of the user. Extensive experiments were conducted on four datasets: GPDS, MCYT, CEDAR and Brazilian PUC-PR datasets. On GPDS-160, we obtained a large improvement in state-of-the-art performance, achieving 1.72\% Equal Error Rate, compared to 6.97\% in the literature. We also verified that the features generalize beyond the GPDS dataset, surpassing the state-of-the-art performance in the other datasets, without requiring the representation to be fine-tuned to each particular dataset.

\end{abstract}


\end{frontmatter}

\fbox{
\parbox[c]{0.9\textwidth}{
\copyright 2017. This manuscript version is made available under the CC-BY-NC-ND 4.0 license \url{http://creativecommons.org/licenses/by-nc-nd/4.0/}
}
}

\section{Introduction}
\label{sec:introduction}

Signature verification systems aim to verify the identity  of individuals by recognizing their handwritten signature. 
They rely on recognizing a specific, well-learned gesture, in order to identify a person. This is in contrast with systems based on the possession of an object (e.g. key, smartcard) or the knowledge of something (e.g. password), and also differ from other biometric systems, such as fingerprint, since the signature remains the most socially and legally accepted means for identification \cite{plamondon_online_2000}.

In offline (static) signature verification, the signature is acquired after the writing process is completed, by scanning a document containing the signature, and representing it as a digital image \cite{impedovo_automatic_2008}. Therefore, the dynamic information about the signature generation process is lost (e.g. position and velocity of the pen over time), which makes the problem very challenging.

\begin{figure*}
\centering
\subfloat[]{\includegraphics[scale=0.42]{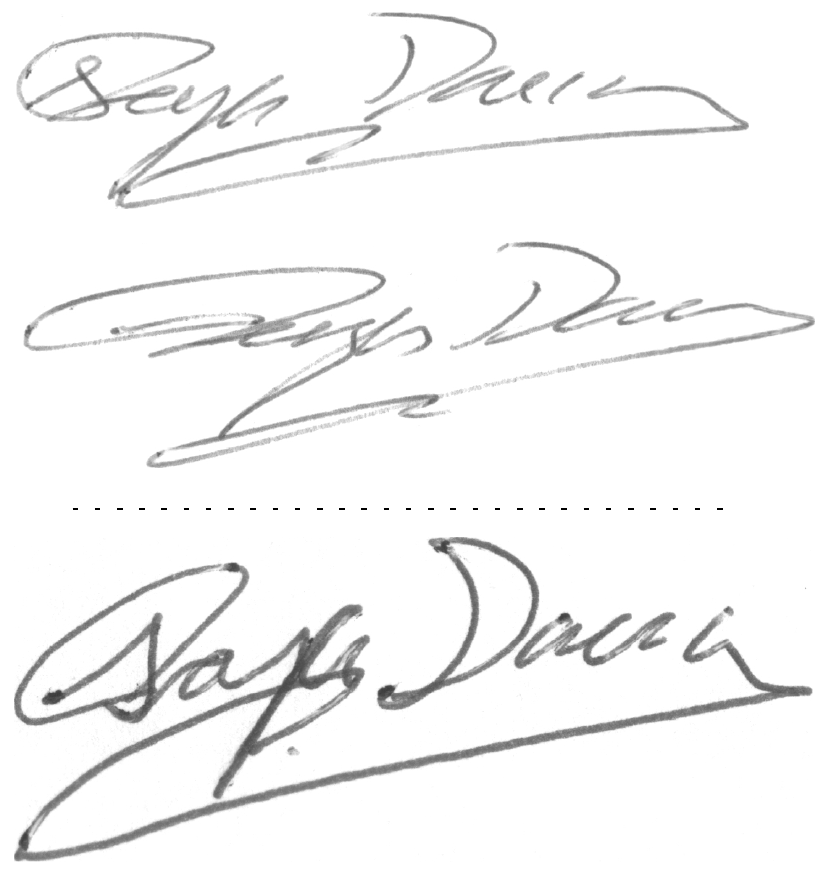}}
\qquad
\subfloat[]{\includegraphics[scale=0.42]{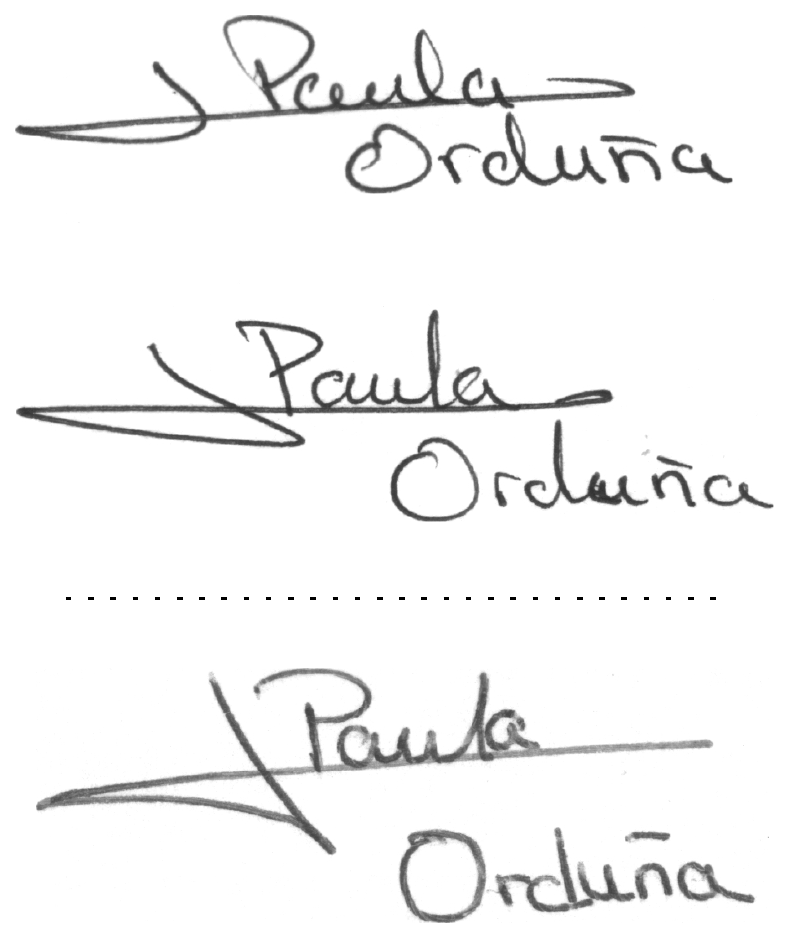}}
\qquad
\subfloat[]{\includegraphics[scale=0.42]{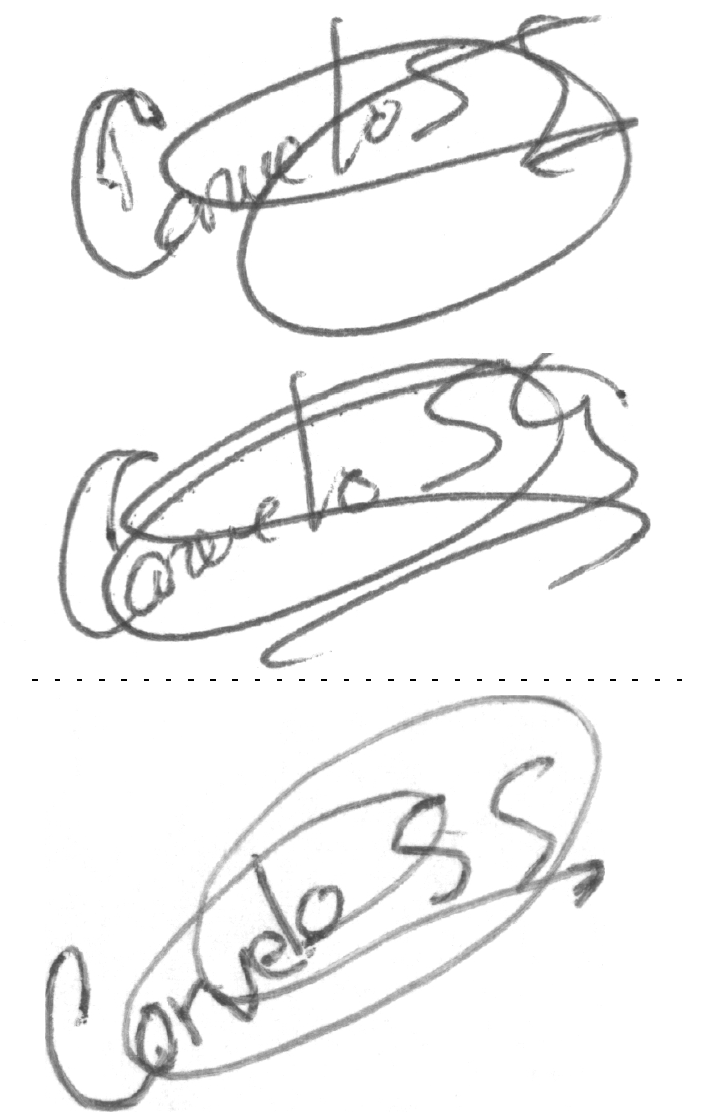}}
\qquad
\subfloat[]{\includegraphics[scale=0.42]{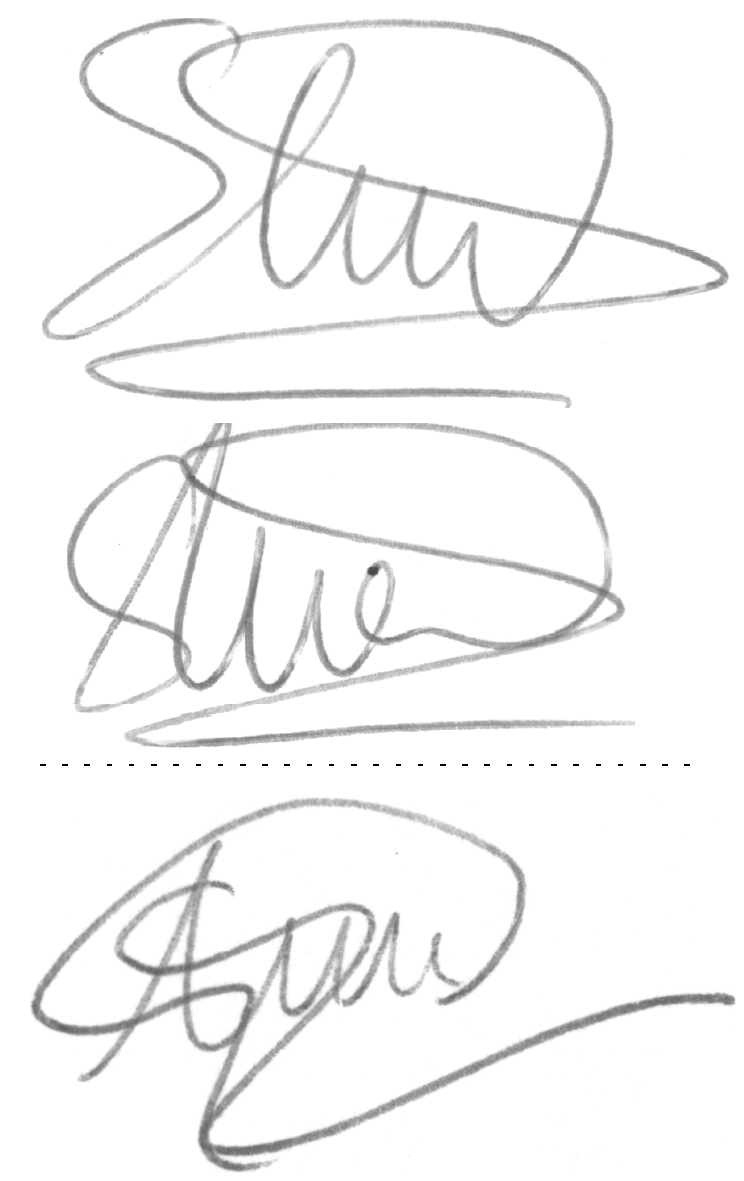}}
\caption{Examples of challenges in designing feature extractors for offline signatures, and the challenge of classifying skilled forgeries. Each column shows two genuine signatures from the same user in the GPDS dataset, and a skilled forgery created for the user. We notice that skilled forgeries resemble genuine signatures to a large extent. Since we do not have examples from the forgery class for training, the problem is even more challenging. We also note the challenges of creating feature extractors for these genuine signatures: \textbf{(a)} The shape of the first name is very different among the two genuine samples. A feature descriptor based on grid features would have very different vectors for the two samples. \textbf{(b)} The shape of the characters in the first name (``Paula'') is very different. An analysis based on the design of individual letters would perform poorly for this user. \textbf{(c)} Large variation in flourishes may impact directional-based descriptors (such as HOG or D-PDF). \textbf{(d)} For some users, it is difficult to pinpoint the common attributes of two signatures even after carefully analyzing the samples.}
\label{fig:signatures}
\end{figure*}

Defining discriminative feature extractors for offline signatures is a hard task. The question ``What characterizes a signature'' is a difficult concept to implement as a feature descriptor, as illustrated in Figure \ref{fig:signatures}. 
This can be observed in the literature, where most of the research efforts on this field have been devoted to finding a good representation for signatures, that is, designing feature extractors tailored for signature verification, as well as using feature extractors created for other purposes \cite{hafemann_offline_2015}.
 Recent work uses texture features, such as Local Binary Patterns (LBP) \cite{yilmaz_score_2016}, \cite{hu_offline_2013} and Gray-Level Co-occurrence Matrix (GLCM) \cite{hu_offline_2013}; directional-based features such as Histogram of Oriented Gradients (HOG) \cite{yilmaz_score_2016} and Directional-PDF \cite{rivard_multi-feature_2013}, \cite{eskander_hybrid_2013}; feature extractors specifically designed for signatures, such as the estimation of strokes by fitting Bezier curves \cite{bertolini_reducing_2010}; among others. No feature extractor has emerged as particularly suitable for signature verification, and most recent work uses a combination of many such techniques.

The difficulty of finding a good representation for signatures reflects on the classification performance of signature verification systems, in particular to distinguish genuine signatures and skilled forgeries - forgeries that are made targeting a particular individual. When we consider experiments conducted on large public datasets, such as GPDS \cite{vargas_off-line_2007}, the best reported results achieve Equal Error Rates around 7\%, even when the number of samples for training is around 10-15, with worse results using fewer samples per user. 

To address both the issue of obtaining a good feature representation for signatures, as well as improving classification performance, we propose a framework for learning the representations directly from the signature images, using convolutional neural networks. 
In particular, we propose a novel formulation of the problem, that incorporates knowledge of skilled forgeries from a subset of users, using a multi-task learning strategy. The hypothesis is that the model can learn visual cues present in the signature images, that are discriminative between genuine signatures and forgeries in general (i.e. not specific to a particular individual). We then evaluate if this feature representation generalizes for other users, for whom we do not have skilled forgeries available.

Our main contributions are as follows: 1) we present formulations to learn features for offline signature verification in a Writer-Independent format. We introduce a novel formulation that uses skilled forgeries from a subset of users to guide the feature learning process, using a multi-task framework to jointly optimize the model to discriminate between users (addressing random forgeries), and to discriminate between genuine signatures and skilled forgeries; 2) we propose a strict experimental protocol, in which all design decisions are made using a validation set composed of a separate set of users. Generalization performance is estimated in a disjoint set of users, from whom we do not use any forgeries for training; 3) we present a visual analysis of the learned representations, which shows that genuine signatures and skilled forgeries get better separated in different parts of the feature space; 4) lastly, we are making two trained models available for the research community\footnote{\url{https://www.etsmtl.ca/Unites-de-recherche/LIVIA/Recherche-et-innovation/Projets}}, so that other researchers can use them as specialized feature extractors for the task.

Experiments were conducted on four datasets, including the largest publicly available signature verification dataset (GPDS), achieving a large performance improvement in the state-of-the-art, reducing Equal Error Rates from 6.97\% to 1.72\% in GPDS-160.  We used the features learned on this dataset to train classifiers for users in the MCYT, CEDAR and Brazilian PUC-PR datasets, also surpassing the state-of-the-art performance, and showing that the learned feature space not only generalizes to other users in the GPDS set, but also to other datasets. 

Preliminary results, using only genuine signatures for learning the features, were published as two conference papers. In \cite{hafemann_ijcnn_2016}, we introduced the formulation to learn features from genuine signatures from a development dataset, using them to train Writer-Dependent classifiers to another set of users. In \cite{hafemann_icpr_2016}, we analyzed the learned feature space and optimized the CNN architecture, obtaining state-of-the-art results on GPDS. The present work includes this formulation of the problem for completeness, with additional experiments on two other datasets (MCYT and CEDAR), a clearer explanation of the method and the experimental protocol, as well as the novel formulation that leverages knowledge of skilled forgeries for feature learning.

The remaining of this paper is organized as follows:
 Section \ref{sec:relatedworks} reviews the related work on signature verification and on feature learning techniques. Section \ref{sec:featurelearning} details the formulation and methodology to learn features for offline signature verification, and section \ref{sec:experimental} describes our experimental protocol. Section \ref{sec:results} presents and discusses the results of our experiments. Lastly, section \ref{sec:conclusion} concludes the paper.

\section{Related works}
\label{sec:relatedworks}

The review of related works is divided below into two parts: we first present a review of previous work on Offline Signature Verification, followed by a brief review of representation learning methods.

\subsection{Related works on Offline Signature Verification}

The area of automatic Offline Signature Verification has been researched at least since the decade of 1970. Over the years, the problem has been addressed from many different perspectives, as summarized by \cite{plamondon_automatic_1989}, \cite{leclerc_automatic_1994} and \cite{impedovo_automatic_2008}. 

In this problem, given a set of genuine signatures, the objective is to learn a model that can distinguish between genuine signatures and forgeries. Forgeries are signatures not created by a claimed individual, and are often subdivided into different types. The most common classification of forgeries in the literature considers: Random Forgeries, where a person uses his or her own signature to impersonate another individual, and Skilled Forgeries, where a person tries to imitate the signature of the claimed individual. While the former is a relatively easier task, discriminating skilled forgeries is an open pattern recognition problem, and is the focus of this paper. This problem is challenging due to a few factors: First, there is a large similarity between genuine signatures and skilled forgeries, as forgers will attempt to imitate the user's signature, often practicing the signature beforehand. Second, in a practical application scenario, we cannot expect to have skilled forgeries for all users in the system, therefore the classifiers should be trained only with genuine signatures in order to be most widely applicable. Lastly, the number of genuine samples per user is often small, especially for new users of the system, for whom we may have only 3 or 5 signatures. This is especially problematic as many users have large intra-class variability, and a few signatures are not sufficient to capture the full range of variation.

There are mainly two approaches for building offline signature verification systems. The most common approach is to design Writer-Dependent classifiers. In this scenario, a training set is constructed for each user of the system, consisting of genuine signatures as positive examples and genuine signatures from other users (random forgeries) as negative samples. A binary classifier is then trained on this dataset, resulting in one model for each user. This approach has shown to work well for the task, but since it requires one model to be trained for each user, complexity increases as more users are enrolled. An alternative is Writer-Independent classification. In this case, a single model is trained for all users, by training a classifier in a dissimilarity space \cite{bertolini_reducing_2010}, \cite{eskander_hybrid_2013}. The inputs for classification are dissimilarity vectors, that represent the difference between the features of a query signature, and the features of a template signature (a genuine signature of the user). In spite of the reduced complexity, Writer-Independent systems often perform worse, and the best results in standard benchmarks are obtained with Writer-Dependent systems.

A large variety of feature extractors have been investigated for this problem, from simple geometric descriptors \cite{nagel_computer_1977}, \cite{justino_off-line_2000}, descriptors inspired in graphology and graphometry \cite{oliveira_graphology_2005}, directional-based descriptors such as HOG \cite{yilmaz_score_2016} and D-PDF \cite{sabourin_off-line_1992}, \cite{rivard_multi-feature_2013}, \cite{eskander_hybrid_2013}, descriptors based on interest-point, such as SIFT \cite{yilmaz_score_2016}, to texture descriptors, such as Local Binary Patterns (LBP) \cite{yilmaz_score_2016} and Gray-Level Co-occurrence Matrix (GLCM) \cite{hu_offline_2013}. These features are commonly extracted locally from the signature images, by dividing the image in a grid and computing descriptors for each cell (either in Cartesian or polar coordinates).

Methods to learn features from data have not yet been widely explored for offline signature verification. Ribeiro et al. \cite{ribeiro_deep_2011} used Restricted Boltzmann Machines (RBMs) to learn features from signature images. However, in this work they only showed the visual appearance of the  weights, and did not test the features for classification. Khalajzadeh \cite{khalajzadeh_persian_2012} used Convolutional Neural Networks (CNNs) for signature verification on a dataset of Persian signatures, but only considered the classification between different users (e.g. detecting random forgeries), and did not considered skilled forgeries. Soleimani et al. \cite{soleimani_deep_2016} proposed a solution using deep neural networks for Multitask Metric Learning. In their work, a distance metric between pairs of signatures is learned. Contrary to our work, the authors used handcrafted feature extractors (LBP in the experiments with the GPDS dataset), while in our work the inputs to the system are the signature themselves (pixel intensities), and the feature representation is learned.
In a similar vein to our work, Eskander \cite{eskander_hybrid_2013} presented a hybrid Writer-Independent Writer-Dependent solution, using a Development dataset for feature selection, followed by training WD classifiers using the selected features. However, in the present work we use a Development dataset for feature learning instead of feature selection.

\subsection{Related work on Representation Learning for computer vision tasks}

In recent years, there has been a large interest in methods that do not rely on hand-crafted features, but rather learn the representations for a problem using \textit{raw} data, such as pixels, in the case of images. Methods based on learning multiple levels of representation have shown to be very effective to process natural data, especially in computer vision and natural language processing \cite{bengio_learning_2009}, \cite{bengio_deep_2013}, \cite{lecun_deep_2015}.
The intuition is to use such methods to learn multiple intermediate representations of the input, in layers, in order to better represent a given problem. In a classification task, the higher layers amplify aspects of the input that are important for classification, while disregarding irrelevant variations \cite{lecun_deep_2015}.
In particular, Convolutional Neural Networks (CNNs) \cite{lecun_backpropagation_1989} have been used to achieve state-of-the-art performance \cite{lecun_deep_2015} in many computer vision tasks \cite{krizhevsky_imagenet_2012}, \cite{szegedy_going_2014}. These models use local connections and shared weights, taking advantage of the spatial correlations of pixels in images by learning and using the same filters in multiple positions of an input image \cite{lecun_deep_2015}. With large datasets, these networks can be trained with a purely supervised criteria. With small datasets, other strategies have been used successfully, such as unsupervised pre-training (e.g. in a greedy layer-wise fashion \cite{bengio_greedy_2006}), and more recently with transfer learning \cite{donahue_decaf:_2013}, \cite{oquab_learning_2014}, \cite{nanni_how_2017}. CNNs have been used to transfer learning of representations, by first training a model in a large dataset, and subsequently using this model in another task (often, a task for which a smaller dataset is available), by using the network as a ``feature extractor'': performing forward-propagation of the samples until one of the last layers before softmax \cite{donahue_decaf:_2013}, \cite{oquab_learning_2014}, or the last layer (that corresponds to the predictions for classes in the original task, as in \cite{nanni_how_2017}), and using the activation at that layer as a feature vector. Alternatively, this pre-trained model can be used to initialize the weights of a model for the task of interest, and training proceeds normally with gradient descent. 

\section{Feature learning for Signature Verification}
\label{sec:featurelearning}

In this work we present formulations for learning features for Offline Signature Verification, and evaluate the performance of such features for training Writer-Dependent classifiers. We first note that a supervised feature learning approach directly applied for Writer-Dependent classification is not practical, since the number of samples per user is very small (commonly around 1-14 samples), while most feature learning algorithms have a large number of parameters (in the order of millions of parameters, for many computer vision problems, such as object recognition \cite{krizhevsky_imagenet_2012}). On the other hand, we expect that signatures from different users share some properties, and we would like to exploit this intuition by learning features across signatures from different writers.

We consider a two-phase approach for the problem: a Writer-Independent feature learning phase followed by Writer-Dependent classification. The central idea is to leverage data from many users to learn a feature space that captures intrinsic properties of handwritten signatures. We subsequently train classifiers for each user, using this feature space, that model the characteristics of each user. Since in real applications the list of users of the system is not fixed, we consider a disjoint set of users for learning the features and training the writer-dependent classifiers, to verify if the learned feature space is useful (i.e. generalizes) to new users. We use the term Writer-Independent for the feature learning process, since the learned representation space is therefore not specific for a set of users.

Given a development set $\mathcal{D}$ of signatures, we train  Deep Convolutional Neural Networks (CNNs) using the formulations defined below. Subsequently, we use the trained network to project the input signatures onto the representation space learned by the CNN for an Exploitation set $\mathcal{E}$, and train a binary classifier for each user. The hypothesis is that genuine signatures and forgeries will be easier to separate in this feature space, if the network succeeds in capturing intrinsic properties of the signatures, that generalizes to other users. 

Convolutional Neural Networks are a particularly suitable architecture for signature verification. This type of architecture scales better than fully connected models for larger input sizes, having a smaller number of trainable parameters. This is a desirable property for the problem at hand, since we cannot reduce the signature images too much without risking losing the details that enable discriminating between skilled forgeries and genuine signatures (e.g. the quality of the pen strokes). We also note that this type of architecture shares some properties with handcrafted feature extractors used in the literature, as features are extracted locally (in an overlapping grid of patches) and combined in non-linear ways (in subsequent layers). In the sections below we present our proposed formulations for the problem, first considering only genuine signatures, and then considering learning from skilled forgeries.

\subsection{Learning features from genuine signatures}
\label{sec:gen}

Let $\mathcal{D}$ be a dataset consisting of genuine signatures from a set of users $\mathcal{Y_D}$. The objective is to learn a function $\phi(X)$ that projects signatures $X$ onto a representation space where signatures and forgeries are better separated.
To address this task, we consider learning a Convolutional Neural Network to discriminate between users in $\mathcal{D}$. This formulation has been introduced in \cite{hafemann_ijcnn_2016}, and it is included here for completeness.

\begin{figure*}
\centering
\includegraphics[width=\textwidth]{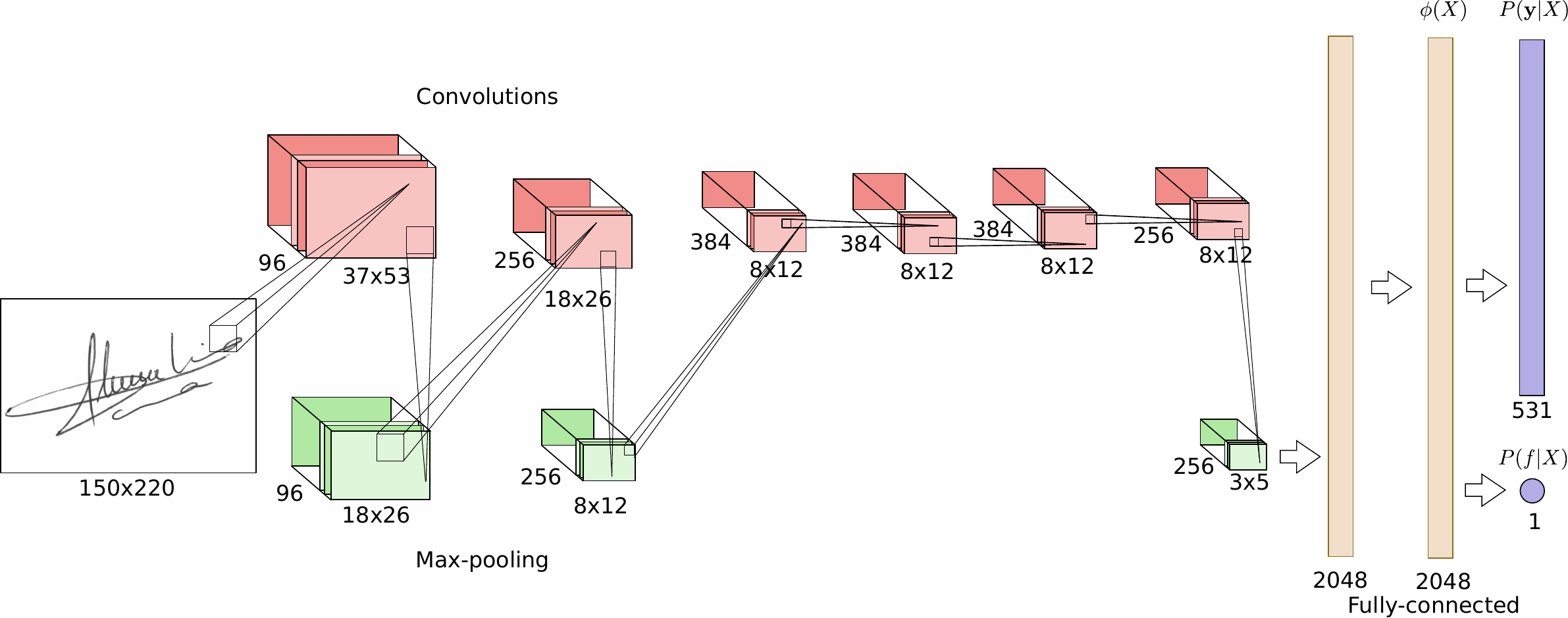}
\caption{Illustration of the CNN architecture used in this work. The input image goes through a sequence of transformations with convolutional layers, max-pooling layers and fully-connected layers. During feature learning, $P(\textbf{y} | X)$ (and also $P(f | X)$ in the formulation from sec \ref{sec:forgneuron}) are estimated by performing forward propagation through the model. The weights are optimized by minimizing one of the loss functions defined in the next sections. For new users of the system, this CNN is used to project the signature images onto another feature space (analogous to ``extract features''), by performing feed-forward propagation until one of the last layers before the final classification layer, obtaining the feature vector $\phi(X)$.}
\label{fig:cnn_architecture}
\end{figure*}

Formally, we consider a training set composed of tuples $(X,y)$ where $X$ is the signature image, and $y$ is the user, that is, $y \in \mathcal{Y_D}$.
We create a neural network with multiple layers, where the objective is to discriminate between the users in the Development set. The last layer of the neural network has $M$ units with a softmax activation, where $M$ is the number of users in the Development set, ($M = | \mathcal{Y_D} |$), and estimates $P(\textbf{y} | X)$. Figure \ref{fig:cnn_architecture} illustrates one of the architectures used in this work, with $M = 531$ users. We train the network to minimize the negative log likelihood of the correct user given the signature image:

\begin{equation}
L = -\sum_j{y_{ij} \log{P(y_{j} | X_i})}
\end{equation}

\noindent Where $y_{ij}$ is the true target for example $i$ ($y_{ij} = 1$ if the signature belongs to user $j$), $X_i$ is the signature image, and $P(y_j | X_i)$ is the probability assigned to class $j$ for the input $X_i$, given by the model. This cost function can then be minimized with a gradient-based method.

The key idea behind this approach is that by training the network to distinguish between users, we expect it to learn a hierarchy of representations, and that the representations on the last layers capture relevant properties of signatures. In particular, if the network succeeds in distinguishing between different users of the development set, then the representation of signatures from these users will be linearly separable in the representation space defined by $\phi(X)$, since the last layer is a linear classifier with respect to its input $\phi(X)$. We test, therefore, the hypothesis that this feature space generalizes well to signatures from other users.

\subsection{Learning features from genuine signatures and skilled forgeries}
\label{sec:forgeries}

One limitation of the formulation above is that there is nothing in the training process to drive the features to be good in distinguishing skilled forgeries. Since this is one of the main goals of a signature verification system, it would be beneficial to incorporate knowledge about skilled forgeries in the feature learning process.

In a real application scenario, we cannot expect to have skilled forgeries available for each user enrolled in the system. We consider, however, a scenario where we obtain skilled forgeries for a subset of the users. Assuming such forgeries are available, we would like to formulate the feature learning process to take advantage of this data. Using the same notation as above, we consider that the development set $\mathcal{D}$ contains genuine signatures and skilled forgeries for a set of users, while the exploitation set $\mathcal{E}$ contains only genuine signatures available for training, and represent the users enrolled to the system.

In this section we introduce novel formulations for the problem, that incorporate forgeries in the feature learning process. The first approach considers the forgeries of each user as a separate class, while the second formulation considers a multi-task learning framework.

\subsubsection{Treat forgeries as separate classes}
\label{sec:forg_as_user}

A simple formulation to incorporate knowledge of skilled forgeries into training is to consider the forgeries of each user as a different class. In this formulation, we have two classes for each user (genuine signatures and forgeries), that is, $M = 2 | \mathcal{Y_D} |$.

We note that this alternative is somewhat extreme, as it considers genuine signatures and forgeries as completely separate entities, while we would expect genuine signatures and skilled forgeries to have a high level of resemblance. 

\subsubsection{Add a separate output for detecting forgeries}
\label{sec:forgneuron}

Another formulation is to consider a multi-task framework, by considering two terms in the cost function for feature learning. The first term drives the model to distinguish between different users (as in the formulations above), while the second term drives the model to distinguish between genuine signatures and skilled forgeries. Formally, we consider another output of the model: $P(f | X)$, a single sigmoid unit, that seeks to predict whether or not the signature is a forgery. The intuition is that in order to classify between genuine signatures and forgeries (regardless of the user), the network will need to learn visual cues that are particular to each class (e.g. bad line quality in the pen strokes, often present in forgeries).

We consider a training dataset containing tuples of the form ($X$, $y$, $f$), where $X$ is the signature image, $y$ is the author of the signature (or the target user, if the signature is a forgery), and $f$ is a binary variable that reflects if the sample is a forgery or not ($f = 1$ indicates a forgery). Note that contrary to the previous formulation, genuine signatures and forgeries targeted to the same user have the same $y$. 
For training the model, we consider a loss function that combines both the classification loss (correctly classifying the user), and a loss on the binary neuron that predicts whether or not the signature is a forgery. The individual losses are shown in Equation \ref{eq:individual_losses}, where the user classification loss ($L_c$) is a multi-class cross-entropy, and the forgery classification ($L_f$) is a binary cross-entropy:

\begin{equation}
\begin{split}
L_c &= -\sum_j{y_{ij} \log{P(y_{j} | X_i})} \\
L_f &= -f_i \log(P(f | X_i)) - (1 - f_i) \log (1 - P(f | X_i))
\end{split}
\label{eq:individual_losses}
\end{equation}

For training the model, we combine the two loss functions and minimize both at the same time. We considered two approaches for combining the losses. The first approach considers a weighted sum of both individual losses:

\begin{equation}
\begin{split}
L_1 &= (1-\lambda) L_c + \lambda L_f \\
&= - (1-\lambda) \sum_j{y_{ij} \log{P(y_{j} | X_i})} + \\ 
&\lambda \big(-f_i \log(P(f | X_i)) - (1 - f_i) \log (1 - P(f | X_i))\big)
\end{split}
\label{eq:model1}
\end{equation}

\noindent Where $\lambda$ is a hyperparameter that trades-off between the two objectives (separating the users in the set $\mathcal{D}$, and detecting forgeries)

In a second approach we consider the user classification loss only for genuine signatures:

\begin{equation}
\begin{split}
L_2 &= (1- f_i) (1-\lambda) L_c + \lambda L_f \\
&= -(1- f_i) (1-\lambda) \sum_j{y_{ij} \log{P(y_{j} | X_i})} + \\
& \lambda \big(-f_i \log(P(f | X_i)) - (1 - f_i) \log (1 - P(f | X_i))\big)
\end{split}
\label{eq:model2}
\end{equation}

In this case, the model is not penalized for misclassifying for which user a forgery was made. 

In both cases, the expectation is that the first term will guide the model to learn features that can distinguish between different users (i.e. detect random forgeries), while the second term will focus on particular characteristics that distinguish between genuine signatures and forgeries (such as limp strokes). It is worth noting that, in the second formulation, using $\lambda = 0$ is equivalent to the formulation in section \ref{sec:gen}, where only genuine signatures are used for training, since the forgeries would not contribute to the loss function.

\subsection{Preprocessing}

The signatures from the datasets used in our experiments are already extracted from the documents where they were written, so signature extraction is not investigated in this paper. Some few preprocessing steps are required, though. The neural networks expect inputs of a fixed size, where signatures vary significantly in shape (in GPDS, they range from small signatures of size 153x258 to large signatures of size 819x1137 pixels). 

We first center the signatures in a large canvas of size $S_\text{canvas} = H \times W$, by using the signatures' center of mass. We remove the background using OTSU's algorithm \cite{otsu_threshold_1975}, setting background pixels to white (intensity $255$), and leaving the foreground pixels in grayscale. The image is then inverted by subtracting each pixel from the maximum brightness  $I(x,y) = 255 - I(x,y)$, such that the background is zero-valued. Lastly, we resize the image to the input size of the network.


\subsection{Training the Convolutional Neural Networks}

For each strategy described above, we learn a feature representation $\phi(.)$ on the Development set of signatures by training a Deep Convolutional Neural Network on this set. This section describes the details of the CNN training.

\begin{table} \centering
\caption{Summary of the CNN layers}
\resizebox{0.6\textwidth}{!}{%
\begin{tabular}{@{}lcc@{}} \hline
Layer & Size & Other Parameters \\ \hline

Input & 1x150x220 \\
Convolution (C1)& 96x11x11 & stride = 4, pad=0 \\
Pooling & 96x3x3 & stride = 2\\

Convolution (C2)& 256x5x5 & stride = 1, pad=2 \\
Pooling & 256x3x3 & stride = 2\\

Convolution (C3)& 384x3x3 & stride = 1, pad=1 \\
Convolution (C4)& 384x3x3 & stride = 1, pad=1 \\
Convolution (C5)& 256x3x3 & stride = 1, pad=1 \\
Pooling & 256x3x3 & stride = 2 \\
Fully Connected (FC6)& 2048 \\
Fully Connected (FC7)& 2048 \\
Fully Connected + Softmax ($P(\textbf{y}|X)$) & M & \\
Fully Connected + Sigmoid ($P(f|X)$) & 1 & \\

\hline
\end{tabular}
}
\label{table:cnn_architecture}
\end{table}

In order to use a suitable architecture for signature verification, in \cite{hafemann_icpr_2016} we investigated different architectures for learning feature representations using the objective from section \ref{sec:gen} (training using only genuine signatures). In this work we use the architecture that performed best for this formulation, which is described in table \ref{table:cnn_architecture}. The CNN consists of multiple layers, considering the following operations: convolutions, max-pooling and dot products (fully-connected layers), where convolutional layers and fully-connected layers have learnable parameters, that are optimized during training. With the exception of the last layer in the network, after each learnable layer we apply Batch Normalization \cite{ioffe2015batch}, followed by the ReLU non-linearity. The last layer uses the softmax non-linearity, which is interpreted as $P(\textbf{y}|X)$ - the probability assigned by the network to each possible user in $\mathcal{Y_D}$. For the formulation in section \ref{sec:forgneuron}, the neuron that estimates $P(f|X)$ uses the sigmoid function. Both output layers receive as input the result of layer FC7. Table \ref{table:operations} lists the operations mentioned above.

Optimization was conducted by minimizing the loss with Stochastic Gradient Descent with Nesterov Momentum, using mini-batches of size $32$, and momentum factor of $0.9$. As regularization, we applied L2 penalty with weight decay $10^{-4}$. The models were trained for 60 epochs, with an initial learning rate of $10^{-3}$, that was divided by $10$ every 20 epochs. We used simple translations as data augmentation, by using random crops of size 150x220 from the 170x242 signature image. As in \cite{ioffe2015batch}, the batch normalization terms (mean and variance) are calculated from the mini-batches during training. For generalization, the mean ($\mathrm{E}[z_i]$) and variance ($\mathrm{Var}[z_i]$) for each neuron were calculated from the entire training set.

\begin{table}
\centering
\resizebox{0.6\textwidth}{!}{%

\begin{threeparttable}
\caption{List of feedforward operations}
\label{table:operations}
\ra{1.5}

\begin{tabular}{lc}
Operation & Formula \\ \hline
Convolution & $\textbf{z}^{l} = \textbf{h}^{l-1} \ast W^{l}$ \\
MaxPooling & $h^{l}_{xy} = \max_{i = 0,..,s ,j = 0,..,s} \textbf{h}^{l-1}_{(x+i)(y+j)} $ \\
Fully-connected layer & $\textbf{z}^l = W^l \textbf{h}^{l-1}$ \\
ReLU & $\text{ReLU}(z_i) = \max(0, z_i)$ \\
Sigmoid & $\sigma(z_i) = \frac{1}{1 + e^{-z_i}}$ \\
Softmax & $\text{softmax}(z_i) = \frac{e^{z_i}}{\sum_{j} e^{z_j}}$ \\
Batch Normalization & $\text{BN}(z_i) = \gamma_i \hat{z_i} + \beta_i$, \\
& $\hat{z_i} = \frac{z_i - \mathrm{E}[z_i]}{\sqrt{\mathrm{Var}[z_i]}}$ \\ \hline
\end{tabular}
\begin{tablenotes}
\footnotesize
\item $\textbf{z}^{l}$: pre-activation output of layer $l$
\item $\textbf{h}^{l}$: activation of layer $l$
\item $\ast$: discrete convolution operator
\item $W$, $\gamma$, $\beta$: learnable parameters
\end{tablenotes}
\end{threeparttable}
}
\end{table}

It is worth noting that, in our experiments, we found Batch Normalization to be crucial to train deeper networks. Without using this technique, we could not train architectures with more than 4 convolutional layers and 2 fully-connected layers. In these cases, the performance in both a training and validation set remained the same as chance, not indicating overfitting, but rather problems in the optimization process.

\subsection{Training Writer-Dependent Classifiers}
\label{sec:wd}

After training the CNN, we use the network to extract feature representations for signatures from the Exploitation set, and train Writer-Dependent classifiers. To do so, we crop the center 150x220 pixels from the 170x242 signature image, perform feedforward propagation until the last layer before softmax (obtaining $\phi(X)$), and use the activations at that layer as the feature vector for the image. This can be seen as a form of transfer learning (similar to \cite{donahue_decaf:_2013}) between the two sets of users. For each user, we build a training set consisting of genuine signatures from the user as positive samples, and genuine signatures from other users as negative samples. We trained Support Vector Machines (SVM), both in a linear formulation and with the Radial Basis Function (RBF) kernel.

We used different weights for the positive and negative class to account for the imbalance of having many more negative samples than positive. The SVM objective becomes \cite{osuna_support_1997}:

\begin{equation}
\begin{aligned}
\min \frac{1}{2}\norm{\textbf{w}}^2& + C^+\Big(\sum_{i:y_i = +1} \xi_i\Big) + C^-\Big(\sum_{i:y_i = -1} \xi_i\Big) \\
\text{subject to} &\\
&y_i(\textbf{w}x_i + b) \geq 1 - \xi_i \\
&\xi_i \geq 0 
\end{aligned}
\end{equation}

\noindent Where the change to the standard SVM formulation is the usage of different weights $C$ for the two classes (we refer the reader to \cite{osuna_support_1997} for the dual formulation). We set the weight of the positive class (genuine signatures) to match the skew (denoted below as $\psi$). Let $P$ be the number of positive (genuine) samples for training, and $N$ the number of negative (random forgery) samples:

\begin{align}
\psi = \frac{N}{P} && C^+ = \psi C^-
\label{eq:c+}
\end{align}

For testing, we used a disjoint set of genuine signatures from the user (that is, not used for training) and the skilled forgeries made targeting the user's signature.

\section{Experimental Protocol}
\label{sec:experimental}

We conducted experiments using the datasets GPDS-960  \cite{vargas_off-line_2007}, MCYT-75 \cite{ortega-garcia_mcyt_2003}, CEDAR \cite{kalera_offline_2004} and the Brazilian PUC-PR \cite{freitas_bases_2000}. Table \ref{table:datasets} summarizes these datasets, including the size used to normalize the images in each dataset (height x width). GPDS-960 is the largest publicly available dataset for offline signature verification with 881 users, having 24 genuine samples and 30 skilled forgeries per user. We used a subset of users from this dataset for learning the features (the development set $\mathcal{D}$) and evaluating how these features generalize to other users in this dataset (the exploitation set $\mathcal{E}$). To enable comparison with previous work, we performed experiments on GPDS having the set $\mathcal{E}$ as the first 160 or the first 300 users of the dataset (to allow comparison with the datasets GPDS-160, and GPDS-300, respectively). In order to evaluate if the features generalize to other datasets, we use the same models learned on GPDS to train Writer-Dependent classifiers for the MCYT, CEDAR and Brazilian PUC-PR datasets.

\begin{table} \centering
\caption{Summary of the datasets used in this work}

\resizebox{\textwidth}{!}{%
\begin{tabular}{@{}ccccc@{}} \hline
Dataset Name  & Users & Genuine signatures & Forgeries & $S_\text{canvas}$ \\ \hline

Brazilian (PUC-PR) & 60 + 108 & 40 & 10 simple, 10 skilled\footnotemark & $700\times1000$  \\ 	
CEDAR  & 55 & 24 & 24 & $730\times1042$ \\	
MCYT-75  &	75	& 15 & 15 & $600\times850$ \\
GPDS Signature 960 Grayscale  & 881 & 24 & 30 & $952\times1360$ \\

\hline
\end{tabular}
}
\label{table:datasets}
\end{table}

\footnotetext{This dataset contains simple and skilled forgeries for the first 60 users}

\begin{figure}
\centering
\includegraphics[width=0.5\textwidth]{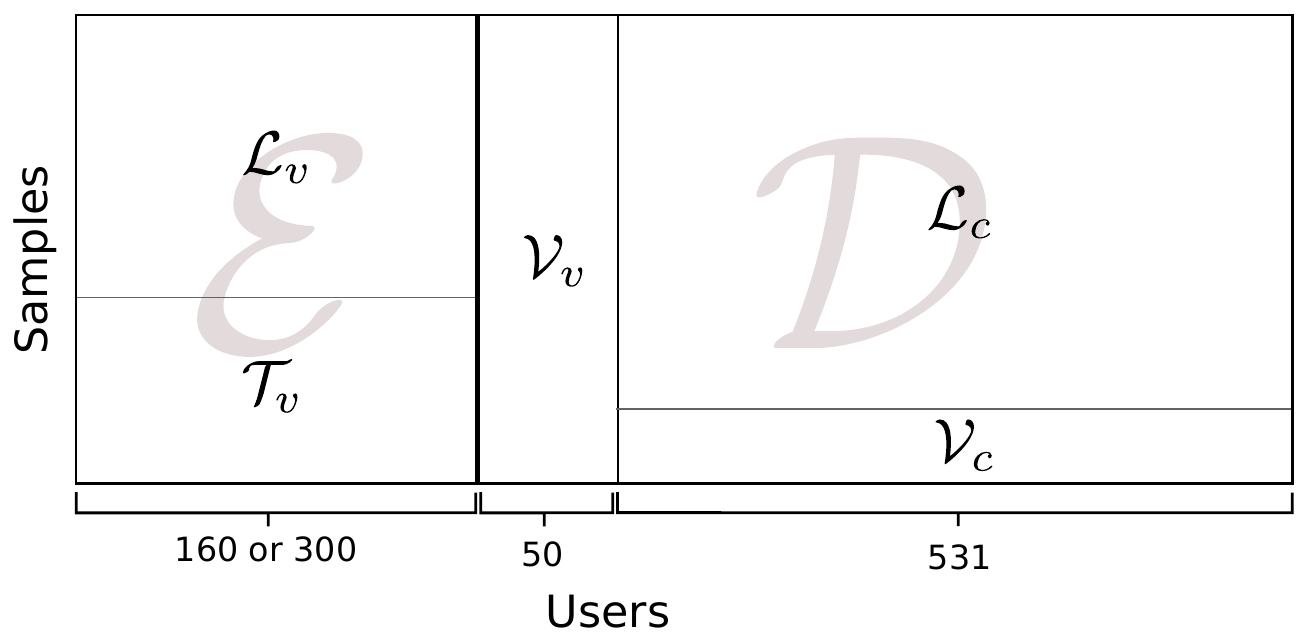}
\caption{The GPDS dataset is separated into an exploitation set $\mathcal{E}$ and Development set $\mathcal{D}$. The development set is used for learning the features, and making all model decisions. The exploitation set represents the users enrolled to the system, where we train Writer-Dependent classifiers using only genuine signatures.}
\label{fig:dataset_separation}
\end{figure}

The GPDS dataset is divided as follows, as illustrated in Figure \ref{fig:dataset_separation}: 
The Convolutional Neural Networks are trained on a set $\mathcal{L}_c$ (denoting \textbf{L}earning set for \textbf{c}lassification) consisting of 531 users. We monitor the progress on a validation set $\mathcal{V}_c$ (\textbf{V}alidation set for \textbf{c}lassification). Both sets contains the same users, but a disjoint set of signature samples from these users. We split 90\% of the signatures for training, and 10\% for this validation set.

After the CNNs are trained, we train Writer-Dependent classifiers on a validation set  $\mathcal{V}_v$ (\textbf{V}alidation set for \textbf{v}erification) consisting of 50 users. The purpose of this set is to allow the estimation of the performance of Writer-Dependent classifiers trained with the representation space learned by the CNN. We use this validation set to make all model choices (CNN architecture and values hyperparameters). On this validation phase, we follow the same protocol for Writer-Dependent classifier training, using a fixed number of 12 genuine signatures for the user as positive samples, and random forgeries from $\mathcal{L}_c$ as negative samples.

Finally, we use the models and hyperparameters that performed best in the validation set, to train and test classifiers for the exploitation set  $\mathcal{E}$. We trained Support Vector Machines on the set $\mathcal{L}_v$  (denoting \textbf{L}earning set for \textbf{v}erification) and tested on $\mathcal{T}_v$ (\textbf{T}esting set for \textbf{v}erification). For each user, we build a dataset consisting of $r$ genuine signatures from the user as positive samples, and genuine signatures from other users as negative samples. Taking into consideration the differences in datasets and experimental protocols that used them in the literature, we used a different number of signatures for training and testing, which is summarized in table \ref{table:datasets_wd}. For the GPDS and the Brazilian PUC-PR datasets, we used signatures from users that are not in the Exploitation set as random forgeries (i.e. signatures from users 301-881 for GPDS-300 and users 61-168 for the Brazilian PUC-PR). For MCYT and CEDAR, we consider genuine samples from other users from the exploitation set as negative samples for training the WD classifier. 
In each experiment, we performed the WD training 10 times, using different splits for the data. We report the mean and variance of the performance across these executions.

\begin{table} \centering
\caption{Separation into training and testing for each dataset}

\resizebox{\textwidth}{!}{%
\begin{tabular}{@{}cccc@{}} \hline
Dataset Name & \multicolumn{2}{c}{Training set} &Testing set\\ 
& Genuine & Random Forgeries & \\ \hline

Brazilian (PUC-PR) & $r \in \{1,\ldots,30\}$ & $30 \times 108 = 3240$ & 10 genuine, 10 random, 10 simple, 10 skilled \\
CEDAR  &  $r \in \{1,\ldots,12\}$ & $12 \times 54 = 648$ & 10 genuine, 10 skilled \\
MCYT-75  & $r \in \{1,\ldots,10\}$ & $10 \times 74 = 588$ & 5 genuine, 15 skilled \\
GPDS-160 & $r \in \{1,\ldots,14\}$ & $14 \times 721 = 10094$ & 10 genuine, 10 random, 10 skilled \\
GPDS-300 & $r \in \{1,\ldots,14\}$ & $14 \times 581 = 8134$ & 10 genuine, 10 random, 10 skilled \\
\hline
\end{tabular}
}
\label{table:datasets_wd}
\end{table}

We used the same hyperparameters for training the SVM classifiers as in previous work \cite{hafemann_icpr_2016}: for the linear SVM, we used $C^-=1$ ($C^+$ is calculated according to equation \ref{eq:c+}). For the SVM with RBF kernel, we used $C^-=1$ and $\gamma = 2^{-11}$.  We found these hyperparameters to work well for the problem, on a range of architectures and users, but we note that they could be further optimized (to each model, or even to each user), which is not explored in this study.

For learning features using forgery data, specifically the formulation on section \ref{sec:forgneuron}, we tested values of $\lambda$ from $0$ to $1$ is steps of $0.1$. The boundaries are special cases: with $\lambda=0$, the forgery neuron is not used at all, and the model only classifies among different users; with $\lambda=1$ the model does no try to separate among different users, but only classifies whether or not the input is a forgery. In our experiments, we found better results on the right end of this range, and therefore we refined the search for the appropriate $\lambda$ with the following cases: $\lambda \in \{0.95, 0.99, 0.999\}$.

Besides comparing the performance with the state-of-the-art in this dataset, we also considered a baseline consisted of a CNN pre-trained on the Imagenet dataset. As argued in \cite{razavian_cnn_2014}, these pre-trained models offer a strong baseline for Computer Vision tasks. We used two pre-trained models\footnote{\url{https://github.com/BVLC/caffe/wiki/Model-Zoo}}, namely Caffenet (Caffe reference network, based on AlexNet \cite{krizhevsky_imagenet_2012}), and VGG-19 \cite{simonyan_very_2014}. We used these networks to extract the feature representations $\phi(X)$ for signatures, and followed the same protocol for training Writing-Dependent classifiers using these representations. We considered the following layers to obtain the representations: pool5, fc6 and fc7.

We evaluate the performance on the testing set using the following metrics: False Rejection Rate (FRR): the fraction of genuine signatures rejected as forgeries; False Acceptance Rate (FAR\textsubscript{random} and FAR\textsubscript{skilled}): the fraction of forgeries accepted as genuine (considering random forgeries and skilled forgeries). We also report the Equal Error Rate (EER): which is the error when FAR = FRR. We considered two forms of calculating the EER: EER\textsubscript{user thresholds}: using user-specific decision thresholds; and EER\textsubscript{global threshold}: using a global decision threshold. In both cases, to calculate the Equal Error Rate we only considered skilled forgeries (not random forgeries) - that is, we use only FRR and FAR\textsubscript{skilled} to estimate the optimum threshold and report the Equal Error Rate. We also report the mean Area Under the Curve (AUC), considering ROC curves created for each user individually. For calculating FAR and FRR in the GPDS exploitation set, we used a decision threshold selected from the validation set $\mathcal{V}_v$ (the threshold that achieved EER using a global decision threshold). 

For the Brazilian PUC-PR dataset, we followed the convention of previous research in this dataset, and also report the individual errors (False Rejection Rate and False Acceptance Rate for different types of forgery) and the Average error rate, calculate as $\text{AER} = (\text{FRR} + \text{FAR\textsubscript{random}} + \text{FAR\textsubscript{simple}} + \text{FAR\textsubscript{skilled}}) / 4$. Since in this work we are mostly interested in the problem of distinguishing genuine signatures and skilled forgeries, we also report $\text{AER\textsubscript{genuine + skilled}} = (\text{FRR} + \text{FAR\textsubscript{skilled}}) / 2$. 

\section{Results and Discussion}
\label{sec:results}

The experimental results with the proposed method are listed and discussed in this section. The first part presents the experiments on the Development set, which was used for making all the design decisions for the proposed method: evaluating different loss functions and other hyperparameters. The second part presents the results on the Exploitation set, and the comparison with the state-of-the-art for all four datasets. 

\subsection{Signature Verification System Design}

In these experiments, we trained the CNN architectures using the loss functions defined in section \ref{sec:featurelearning}, used them to extract features for the users in the validation set $\mathcal{V}_v$, and trained Writer-Dependent classifiers for these users using 12 reference signatures. We then analyzed the impact in classification performance of the different formulations of the problem.

\begin{figure*}
\centering
\subfloat[Loss $L_1$, Linear SVM]{{\includegraphics[width=0.45\textwidth]{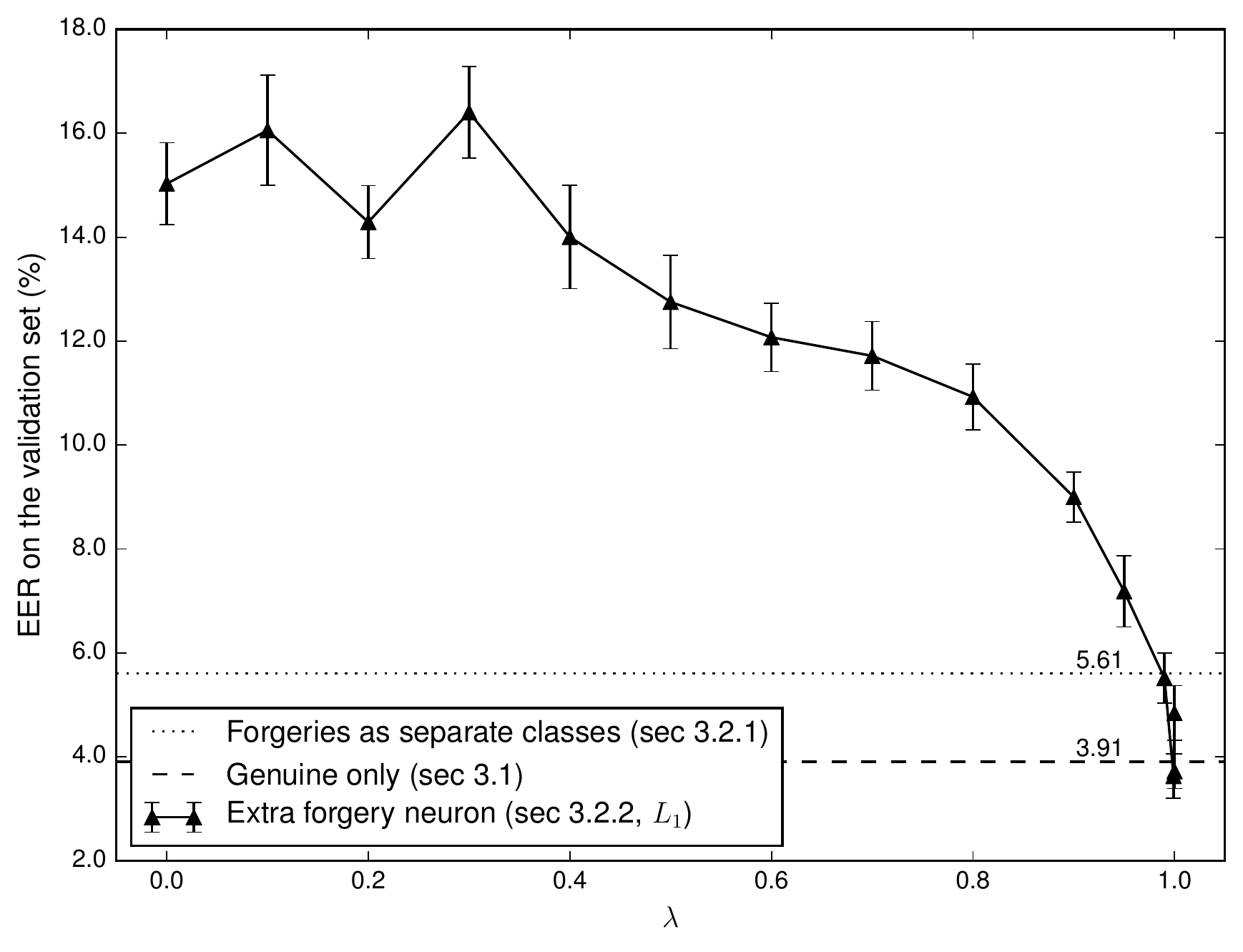} }}
\qquad
\subfloat[Loss $L_1$, SVM with RBF kernel]{{\includegraphics[width=0.45\textwidth]{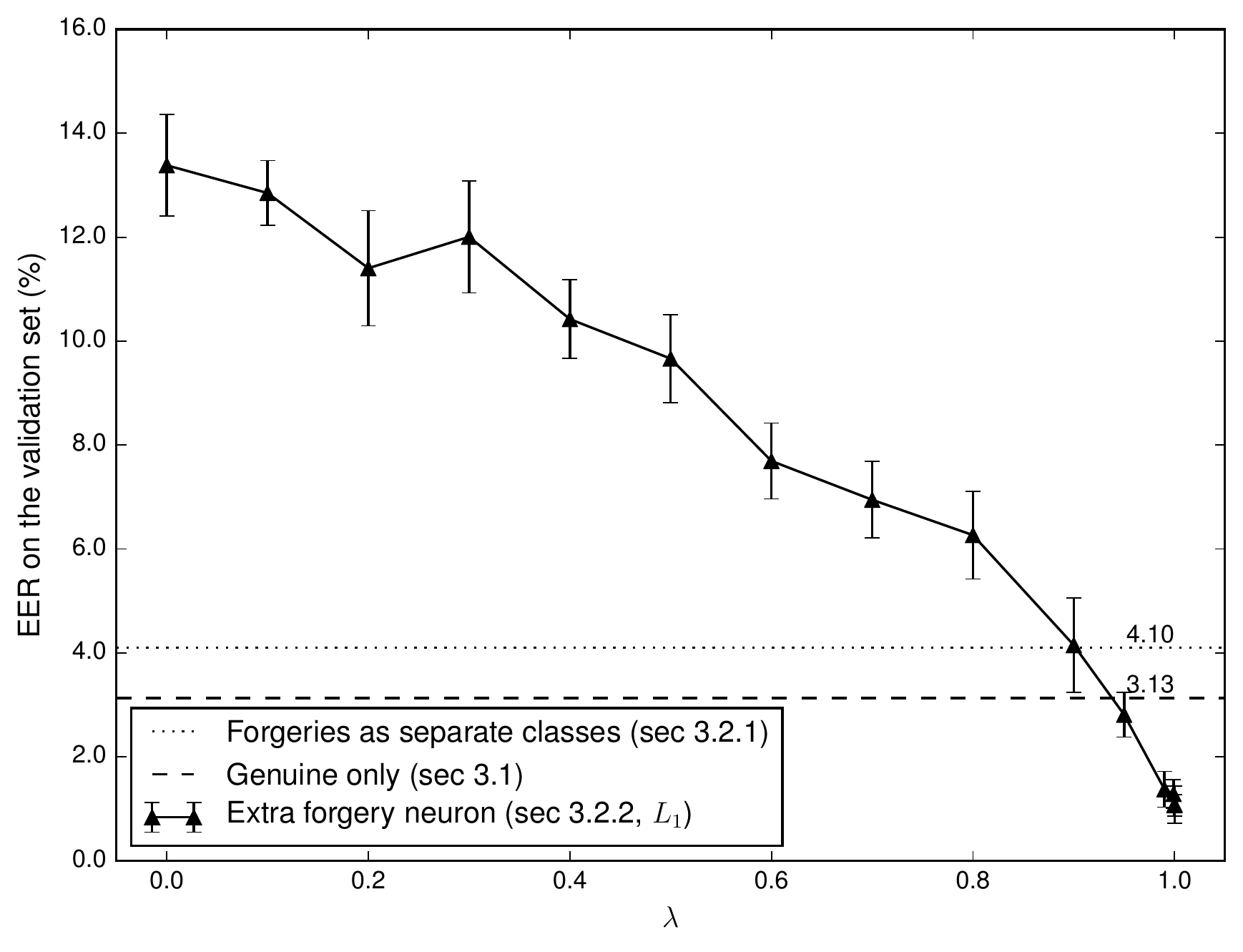} }}
\qquad
\subfloat[Loss $L_2$, Linear SVM]{{\includegraphics[width=0.45\textwidth]{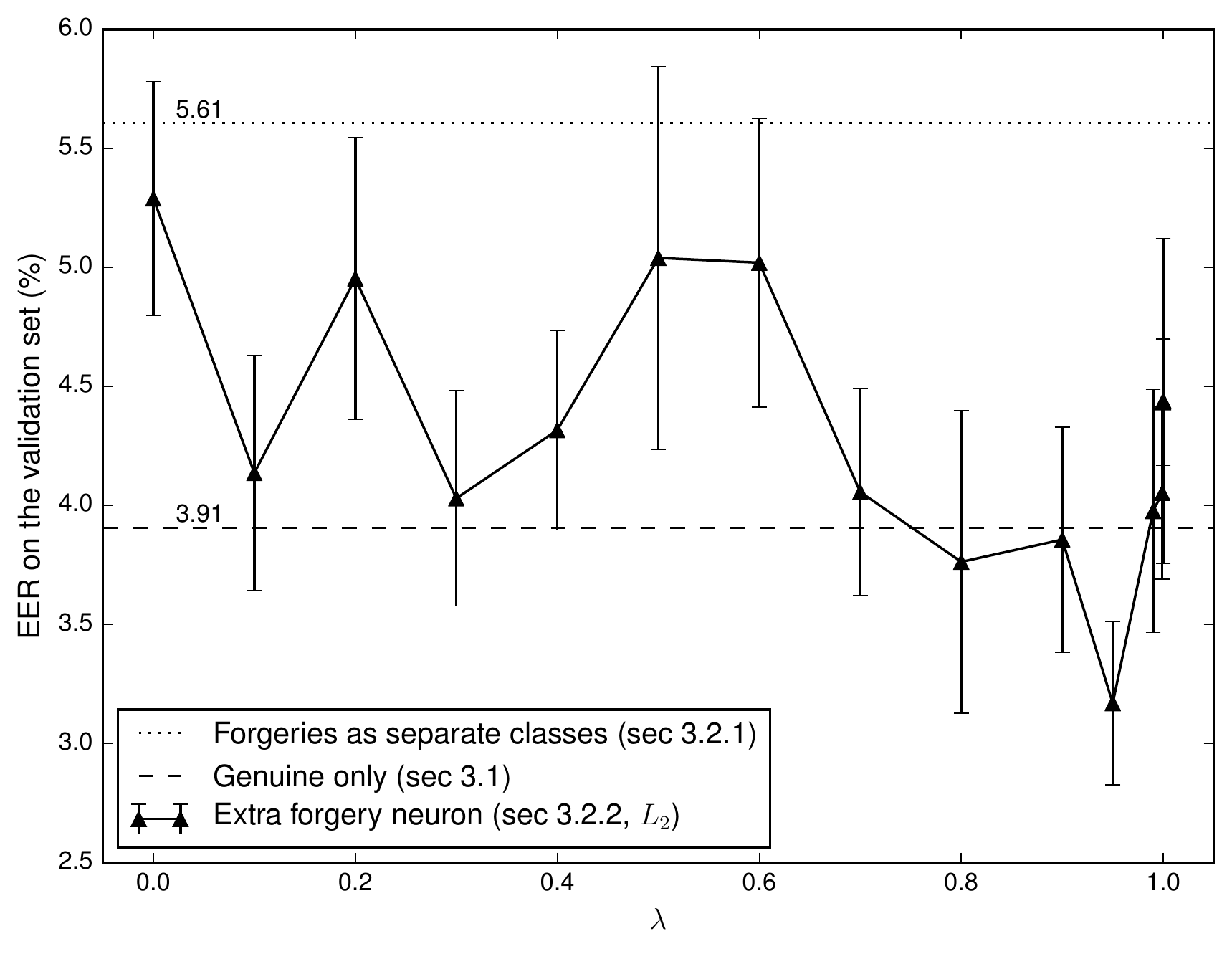} }}
\qquad
\subfloat[Loss $L_2$, SVM with RBF kernel]{{\includegraphics[width=0.45\textwidth]{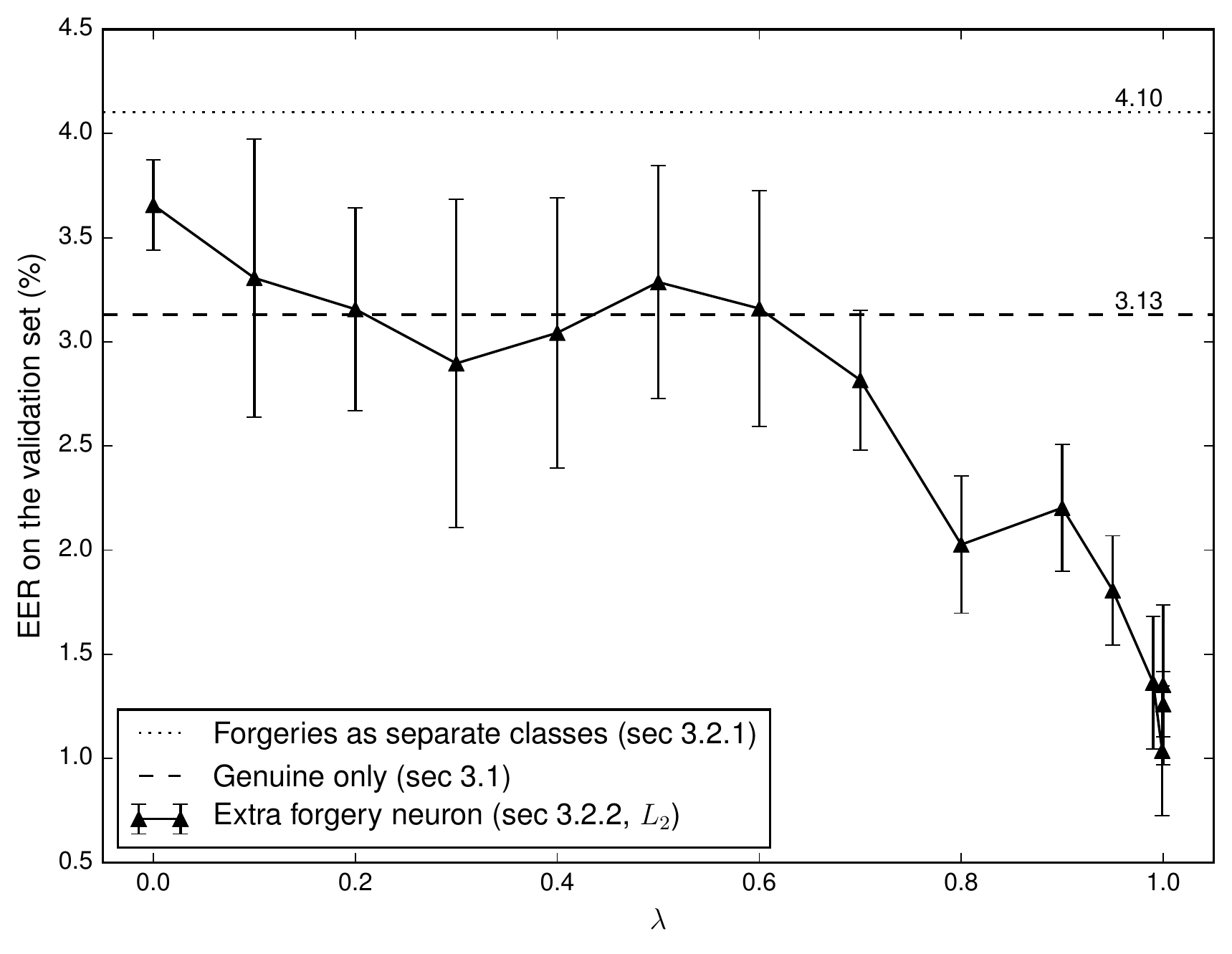} }}
\caption{Performance on the validation set ($\mathcal{V}_v$), using features learned from genuine signatures and forgeries (sec \ref{sec:forgneuron}), as we vary the hyperparameter $\lambda$. For reference, the performance of models using features learned from genuine signatures only (sec \ref{sec:gen}) and using forgeries as different classes (sec \ref{sec:forg_as_user}) are also included.}
\label{fig:varying_lambda}
\end{figure*}

For the formulation on section \ref{sec:forgneuron}, where we have a separate neuron to estimate if a signature is a forgery or not, we trained models with variable values of $\lambda$. Figure \ref{fig:varying_lambda} shows the results on the validation set using loss $L_1$ (from equation \ref{eq:model1}), and loss $L_2$ (from equation \ref{eq:model2}). The models with loss $L_2$ only consider the user-classification loss for genuine signatures, while the models using $L_1$ consider user-classification loss for all signatures (genuine and forgeries). As a performance reference, we also show the results using a model trained with genuine signatures only, as well as the model trained with forgeries as separate classes (sec \ref{sec:forg_as_user}).

Both using a linear SVM or using an SVM with RBF kernel, the results using the loss $L_1$ were very poor for low values of $\lambda$. This is likely caused by the fact that, in this formulation, both genuine signatures and forgeries of the same user are assigned to the same class $y$, and the loss function guides the model to be less discriminative between the genuine signatures and forgeries of the same user. This behavior is not present when we use the loss $L_2$, since the model is not penalized for misclassifying for which user the forgery was created. We also noticed that the best results were closer to the right end of the range, suggesting that the distinction of forgeries (regardless of the user) in the development set may be more relevant than the distinguishing genuine signatures from different users. In the extreme case, with $\lambda = 1$, the model is only learning to discriminate between genuine signatures and forgeries (the output is a single binary unit), and the performance is still reasonable, although worse than the performance when both loss functions are combined. 
It is worth noting that the scale of $L_c$ is larger than $L_f$ by definition: $L_c$ is a cross-entropy loss among 531 users. A random classifier would have loss $L_c  \approx \log(531) \approx 6.27$. On the other hand, $L_f$ is a cross-entropy loss among 2 alternatives, and a random classifier would have loss around $L_f \approx \log(2) \approx 0.69$, which also partially explains larger $\lambda$ values.

We noticed an unexpected behavior using loss $L_2$ with $\lambda=0$. This loss function is equivalent to the loss when using only genuine signatures, but actually performed worse during our experiments. Analyzing this abnormal behavior, we identified that, although the forgeries do not contribute to the loss function directly, they do have some indirect effect on loss function due to the usage of batch normalization. During training, the skilled forgeries are used, together with genuine signatures, when computing the batch statistics (mean and variance), therefore affecting the output of the network. However, it is unclear why this effect results in worse performance, instead of simply adding more variance to the results.

We also verified if the forgery neuron generalized well to other users. Since this neuron is not related to a particular user in the development set, we can use it to estimate $P(f|X)$ for signature images from other users. In this case, we estimate if a signature is a forgery only by looking at the questioned specimen, and not comparing it to other genuine signatures from the same user. We used the neuron trained with loss $L_2$ and $\lambda = 0.999$ to classify all signatures from the validation set $\mathcal{V}_v$, achieving an error rate of 14.37\%. In comparison, for classifying signatures from the same set of users where the CNN was trained (i.e. testing on $\mathcal{V}_c$), the model achieved 2.21\% of error. This suggests that using this neuron is mostly helpful to guide the system to obtain better representations (and subsequently train WD classifiers), than to use it directly as a classifier for new samples, since it mainly generalizes to other signatures from the same users used to train the CNN.

\begin{table*}
\centering
\caption{Performance of the WD classifiers on the validation set $\mathcal{V}_v$ (subset of 50 users in GPDS; Errors and Standard deviations in \%)} 
\label{tbl:perf_validation}
\resizebox{\textwidth}{!}{%
\ra{1.5}
\begin{tabular}{lllll}
\hline
Classifier &                                              Formulation used to learn the features  & EER\textsubscript{global threshold} & EER\textsubscript{user thresholds} &             Mean AUC \\
\hline
Linear SVM &                   Baseline (Caffenet, layer pool5) &     14.09 (+- 2.80) &    10.59 (+- 2.96) &  0.9453 (+- 0.0198) \\
 &                  Using genuine signatures only (sec \ref{sec:gen})&      6.80 (+- 0.57) &     3.91 (+- 0.64) &  0.9876 (+- 0.0022) \\
 &     Forgeries as separate classes (sec \ref{sec:forg_as_user}) &      9.45 (+- 0.51) &     5.61 (+- 0.63) &  0.9749 (+- 0.0028) \\
 &  Forgery neuron (sec \ref{sec:forgneuron}, loss $L_1$, $\lambda = 0.999$) &      7.01 (+- 0.42) &     3.63 (+- 0.43) &  0.9844 (+- 0.0024) \\
 &   Forgery neuron (sec \ref{sec:forgneuron}, loss $L_2$, $\lambda = 0.95$) &      6.09 (+- 0.29) &     3.17 (+- 0.34) &  0.9899 (+- 0.0017) \\
\hline

SVM (RBF) &                                        Baseline (Caffenet, layer fc6) &     16.20 (+- 0.94) &    13.51 (+- 0.99) &  0.9261 (+- 0.0054) \\
 &                                     Using genuine signatures only (sec \ref{sec:gen}) &      5.93 (+- 0.43) &     3.13 (+- 0.46) &  0.9903 (+- 0.0018) \\
 &                        Forgeries as separate classes (sec \ref{sec:forg_as_user}) &      7.79 (+- 0.43) &     4.10 (+- 0.41) &  0.9857 (+- 0.0012) \\
 &     Forgery neuron (sec \ref{sec:forgneuron}, loss L1, $\lambda = 1$) &      2.41 (+- 0.32) &     1.08 (+- 0.36) &  0.9978 (+- 0.0008) \\
 &  Forgery neuron (sec \ref{sec:forgneuron}, loss L2, $\lambda = 0.999$) &      2.51 (+- 0.33) &     1.04 (+- 0.31) &  0.9971 (+- 0.0009) \\
\hline

\end{tabular}
} 
\end{table*}

Table \ref{tbl:perf_validation} consolidates the performance obtained in the validation set $\mathcal{V}_v$ using the proposed methods. The baseline, using a CNN pre-trained on the ImageNet dataset, performed reasonably well compared to previous work on the GPDS dataset, but still much worse than the methods that learned on signature data. 
An interesting result is that the naive formulation to use forgeries (treat forgeries as separate classes - section \ref{sec:forg_as_user}) performed worse than the formulation that used only genuine signatures for training the CNN. Using the model trained with genuine signatures, we obtained EER of 3.91\% using a linear SVM, and 3.13\% using the RBF kernel. Using the model trained with forgeries as separate classes, we obtained EER of 5.61\% using Linear SVM and 4.10\% using the RBF kernel. A possible explanation for this effect is that this formulation effectively doubles the number of classes, making the classification problem much harder. This fact, combined with the observation that genuine signatures and forgeries for the same user usually share several characteristics, may justify this drop in performance. On the other hand, the formulation using the forgery neuron performed much better in the validation set, showing that this is a promising formulation of the problem. We reiterate that forgeries are used only in the feature learning process, and that no forgeries from the validation set $\mathcal{V}_v$ were used for training.

Although it is not the focus of this paper, we note that these models could also be used for user identification from signatures. Using the features learned from genuine signatures only (sec \ref{sec:gen}), the performance on the validation set $\mathcal{V}_c$ (classification between the 531 users) is 99.23\%, showing that using CNNs for this task is very effective.

\subsubsection{Visualizing the learned representation space}

We performed an analysis of the feature space learned by the models, by using the t-SNE algorithm \cite{van2008visualizing} to project the samples from the validation set $\mathcal{V}_v$ from $\mathbb{R}^N$ to $\mathbb{R}^{2}$. This analysis is useful to examine the local structure present in this high-dimensionality space. For this analysis, we used the baseline model (Caffenet, using features from layer pool5), a model learned with genuine signatures only, and a model learned with genuine signatures and forgeries (using loss $L_2$ and $\lambda = 0.95$). These models were trained on the set $\mathcal{L}_c$, which is a disjoint set of users from the validation set. In all cases, we used the models to ``extract features'' from all 1200 signatures images from the validation set, by performing forward propagation until the layer specified above. For the baseline model, this representation is in $\mathbb{R}^{9216}$, while for the other models it is in $\mathbb{R}^{2048}$. For each model, we used the t-SNE algorithm to project the samples to 2 dimensions.

\begin{figure*}
\centering
\subfloat[Baseline (Features learned on Imagenet)]{{\includegraphics[width=0.29\textwidth]{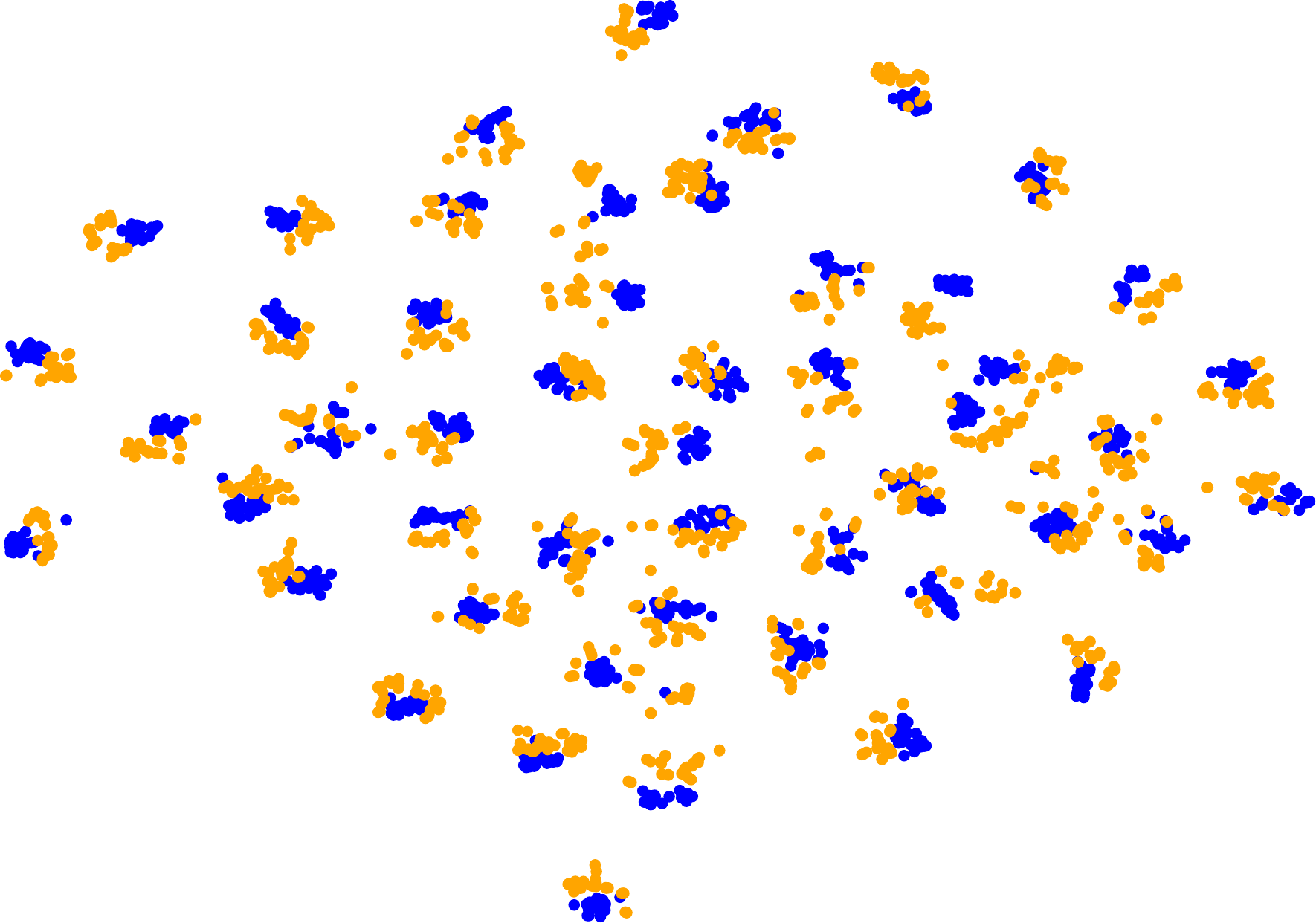} }}
\qquad
\subfloat[Using only genuine signatures to learn the features]{{\includegraphics[width=0.29\textwidth]{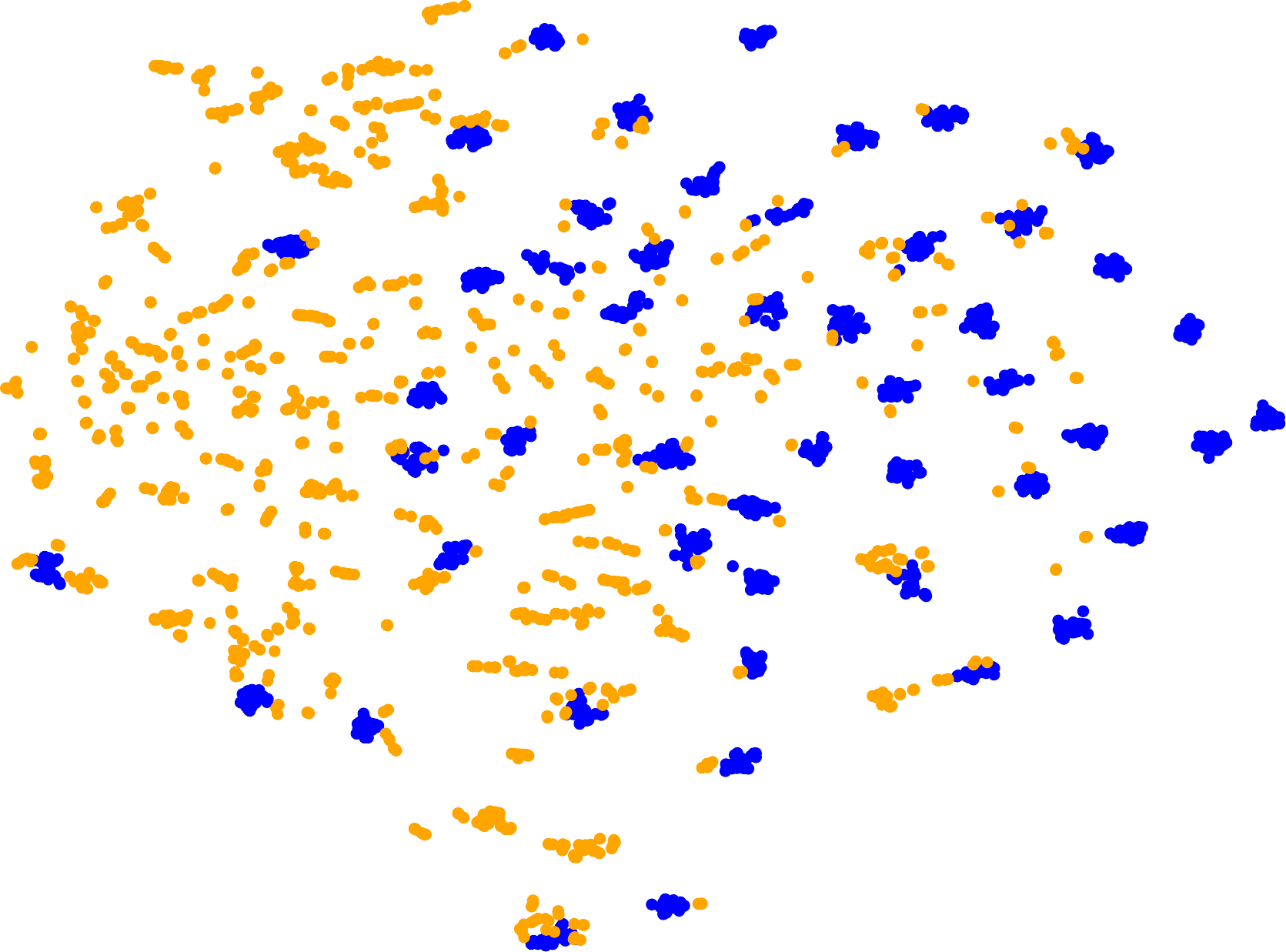} }}
\qquad
\subfloat[Using genuine signatures and forgeries to learn the features]{{\includegraphics[width=0.29\textwidth]{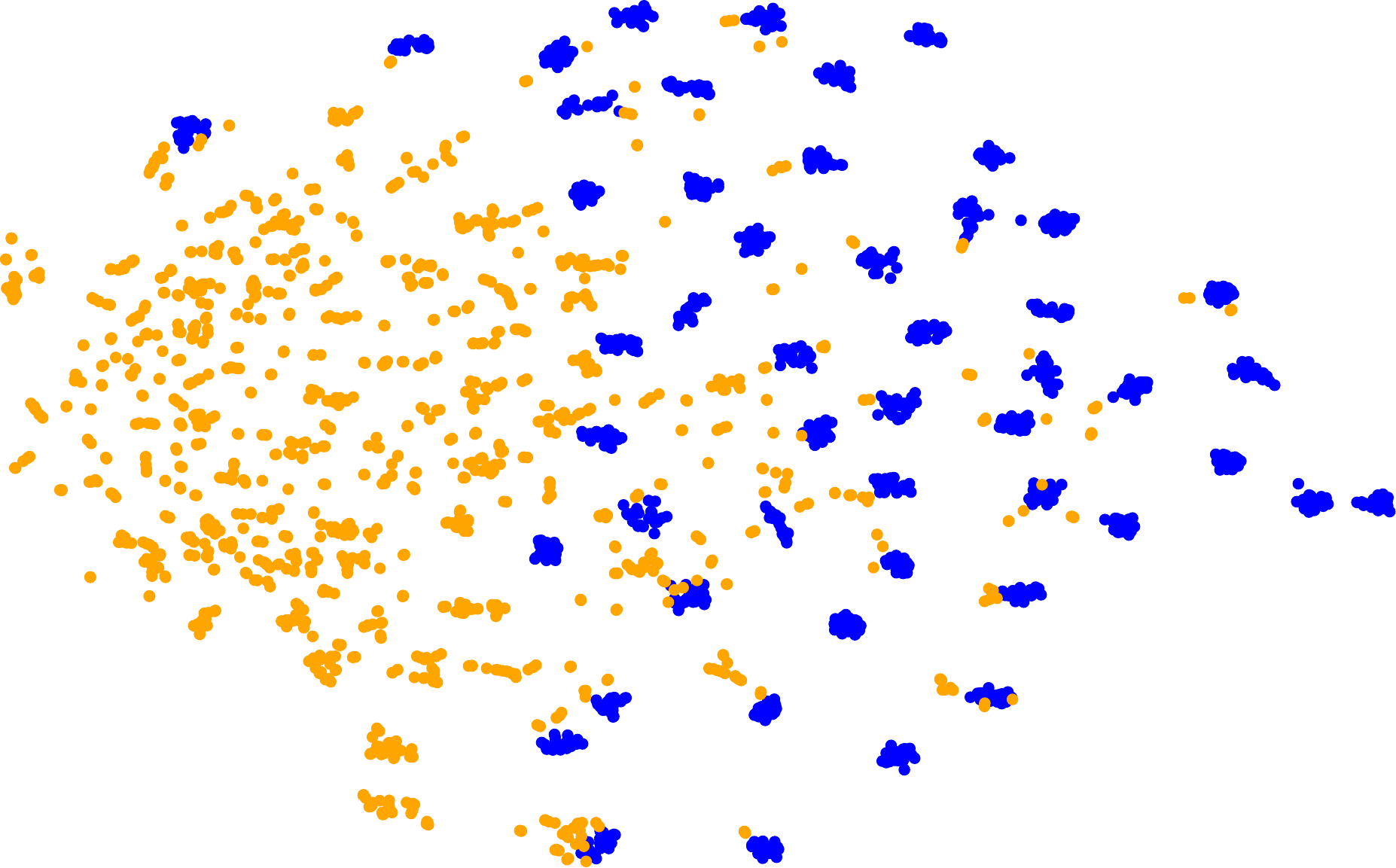} }}
\caption{2D projections (using t-SNE) of the feature vectors from the 50 users in the validation set $\mathcal{V}_v$. Each point represents a signature sample: genuine signatures are displayed in blue (dark), while skilled forgeries are displayed in orange (light).}
\label{fig:tsne_grayscale}
\end{figure*}

The result can be seen in Figure \ref{fig:tsne_grayscale}. The baseline system (model trained on natural images) projects the samples onto a space where samples from different users are clustered in separate regions of the space, which is is quite interesting considering that this network was never presented signature images. On the other hand, skilled forgeries are also clustered together with genuine signatures in this representation. On the models trained with signature data, we can see that signatures from different users also occupy different regions of the feature space. Using the model trained with genuine signatures and forgeries, we see that the forgeries from the users in the validation set are much more grouped together in a part of the feature space, although several forgeries are still close to the genuine signatures of the users. This suggests that the network has learned characteristics that are intrinsic to many forgeries, that generalizes to other users.

\subsection{Generalization performance and comparison with the state-of-the-art}

We now present the results on the exploitation set, comparing the results with the state-of-the-art. In these experiments, we do not use any skilled forgeries from the users, since it is not reasonable to expect skilled forgeries to be available for all users enrolled in the system.

We reiterate that all design decisions (e.g. choice of architecture and other hyperparameters) were done using the validation set $\mathcal{V}_v$, which consists of a separate set of users, to present an unbiased estimate of the performance of the classifier in the testing set. In these experiments, we used the architectures that performed best in the validation set, as seen in Table \ref{tbl:perf_validation}. In particular, we consider a model that was learned using genuine signatures only (sec \ref{sec:gen}), which we call simply by \textbf{SigNet} in this section. We also consider a model learned using genuine signatures and forgeries (sec \ref{sec:forgneuron}), using loss $L2$, which we call \textbf{SigNet-F}. For the experiments with a linear SVM, we used the model learned with $\lambda = 0.95$, while for the experiments with the SVM with the RBF kernel, we used the model learned with $\lambda = 0.999$.

\subsubsection{Experiments on GPDS-160 and GPDS-300}

For these experiments, we used the models SigNet and SigNet-F to extract features of the exploitation set (GPDS-160 and GPDS-300), and trained Writer-Dependent classifiers. To report the False Rejection Rate and False Acceptance Rates, we used the validation set to find the optimum global threshold (the threshold that obtained EER\textsubscript{global threshold} on the validation set $\mathcal{V}_v$) as a global threshold for all users. In this work, we do not explore techniques for setting user-specific thresholds, but simply report EER\textsubscript{user thresholds}, which is the equal error rate obtained by using the optimal decision thresholds for each user.

\begin{table*}
\centering
\caption{Detailed performance of the WD classifiers on the GPDS-160 and GPDS-300 datasets (Errors and Standard Deviations in \%)}
\label{tbl:gpds_detail}
\resizebox{\textwidth}{!}{%
\ra{1.5}
\begin{tabular}{lrlllllll}
\hline
  Dataset &  Samples per user &    Classifier &             FRR &      FAR\_random &     FAR\_skilled & EER\textsubscript{global threshold} & EER\textsubscript{user thresholds} &             meanAUC \\
\hline
GPDS-160 &                 5 &  SVM (Linear) &  9.09 (+- 0.65) &  0.01 (+- 0.03) &  5.75 (+- 0.12) &      7.30 (+- 0.35) &     3.52 (+- 0.28) &  0.9880 (+- 0.0013) \\
 &                  &     SVM (RBF) &  5.16 (+- 0.41) &  0.06 (+- 0.04) &  5.17 (+- 0.17) &      5.15 (+- 0.22) &     2.41 (+- 0.12) &  0.9924 (+- 0.0011) \\
&                12 &  SVM (Linear) &  6.39 (+- 0.67) &  0.01 (+- 0.02) &  3.96 (+- 0.18) &      5.15 (+- 0.28) &     2.60 (+- 0.39) &  0.9922 (+- 0.0010) \\
&                &     SVM (RBF) &  3.59 (+- 0.23) &  0.02 (+- 0.03) &  3.66 (+- 0.15) &      3.61 (+- 0.07) &     1.72 (+- 0.15) &  0.9952 (+- 0.0006) \\
 GPDS-300 &                 5 &  SVM (Linear) &  9.28 (+- 0.36) &  0.01 (+- 0.02) &  8.18 (+- 0.23) &      8.68 (+- 0.22) &     4.84 (+- 0.26) &  0.9792 (+- 0.0016) \\
&                  &     SVM (RBF) &  6.03 (+- 0.45) &  0.04 (+- 0.04) &  4.68 (+- 0.18) &      5.25 (+- 0.15) &     2.42 (+- 0.24) &  0.9923 (+- 0.0007) \\
&                12 &  SVM (Linear) &  6.80 (+- 0.31) &  0.00 (+- 0.01) &  6.16 (+- 0.17) &      6.44 (+- 0.17) &     3.56 (+- 0.18) &  0.9857 (+- 0.0010) \\
&                &     SVM (RBF) &  3.94 (+- 0.29) &  0.02 (+- 0.02) &  3.53 (+- 0.11) &      3.74 (+- 0.15) &     1.69 (+- 0.18) &  0.9951 (+- 0.0004) \\
\hline
\end{tabular}
}
\end{table*}

Table \ref{tbl:gpds_detail} lists the detailed results on the GPDS-160 and GPDS-300 datasets, for experiments using SigNet-F. We notice that the using only 5 samples per user already achieves a good average performance on these datasets, showing that the proposed strategy works well with low number of samples per user. We also note that the performance using user-specific thresholds is much better than using a single global threshold (1.72\% vs 3.61\%) in the GPDS-160 dataset, which is consistent with previous findings that the definition of user-specific thresholds is key in obtaining a good performance.

We notice that the performance using a linear classifier (Linear SVM) is already good, which is interesting from a practical perspective for a large-scale deployment. Since the CNN model is the same for all users, adding new users to the system requires only training the WD classifier. For a linear classifier, this requires only one weight per dimension (plus a bias term), adding to 2049 doubles to be stored (16KB per user). For the SVM with RBF kernel, the storage requirements for each user depends on the number of support vectors. In the GPDS-300 dataset, in average the classifiers used 75 support vectors. Since the set of random forgeries is the same for all users, most of these support vectors will be shared among different users. On the other hand, we noticed that the majority of genuine signatures were selected as support vectors (as expected) - in average 10.3 genuine signatures, when using 12 references for training.

\begin{table}
\centering
\caption{Comparison with state-of-the art on the GPDS dataset (errors in \%)}
\label{tbl:soa_gpds}

\resizebox{0.6\textwidth}{!}{%
\begin{threeparttable}
\ra{1.5}

\begin{tabular}{llllr}
\hline
 Reference & Dataset& \begin{tabular}[x]{@{}c@{}}\#samples\\per user\end{tabular}  &Features &  EER\\
\hline
Hu and Chen \cite{hu_offline_2013}& GPDS-150 &10 &LBP, GLCM, HOG & 7.66\\
Guerbai et al \cite{guerbai_effective_2015} & GPDS-160&12&Curvelet transform&15.07\\
Serdouk et al \cite{serdouk_new_2015} & GPDS-100 & 16 & GLBP, LRF& 12.52 \\
Yilmaz \cite{yilmaz_score_2016}  &GPDS-160 &5&LBP, HOG, SIFT& 7.98\\
Yilmaz \cite{yilmaz_score_2016}  &GPDS-160 &12&LBP, HOG, SIFT& 6.97\\
Soleimani et al \cite{soleimani_deep_2016}  &GPDS-300 &10&LBP& 20.94\\

\hline

Present Work  & GPDS-160 & 5  & SigNet & 3.23 (+-0.36)\\
Present Work  & GPDS-160 & 12  & SigNet & 2.63 (+-0.36)\\
Present Work  & GPDS-300 & 5  & SigNet & 3.92 (+-0.18)\\
Present Work  & GPDS-300 & 12  & SigNet & 3.15 (+-0.18)\\

Present Work  & GPDS-160 & 5  & SigNet-F & 2.41 (+-0.12)\\

\textbf{Present Work } & \textbf{GPDS-160} & \textbf{12}  & \textbf{SigNet-F}& \textbf{1.72 (+-0.15)}\\

Present Work  & GPDS-300 & 5  & SigNet-F  & 2.42 (+-0.24)\\

\textbf{Present Work}  & \textbf{GPDS-300} & \textbf{12}  & \textbf{SigNet-F}& \textbf{1.69 (+-0.18)}\\

\hline
\end{tabular}

\end{threeparttable}
}
\end{table}

Table \ref{tbl:soa_gpds} compares our results with the state-of-the-art on the GPDS dataset. We observed a large improvement in verification performance, obtaining 1.72\% EER on GPDS-160, compared to a state-of-the-art of 6.97\%, both using 12 samples per user for training. We also note that this result is obtained with a single classifier, while the best results in the state-of-the-art use ensembles of many classifiers. As in the experiments in the validation set, we notice an improvement in performance using SigNet-F to extract the features compared to using SigNet.

\subsubsection{Generalizing to other datasets}

We now consider the generalization performance of the features learned in GPDS to other datasets. We use the same networks, namely SigNet and SigNet-F, for extracting features and training Writer-Dependent classifiers on MCYT, CEDAR and the Brazilian PUC-PR datasets.

\begin{table}
\centering
\caption{Comparison with the state-of-the-art in MCYT}
\label{tbl:mcyt_soa}
\resizebox{0.6\textwidth}{!}{%
\begin{tabular}{lrlc}
\hline
Reference & \# Samples & Features & EER\\
\hline
Gilperez et al.\cite{gilperez_off-line_2008}&5&Contours (chi squared distance)&10.18\\
Gilperez et al.\cite{gilperez_off-line_2008}&10&Contours (chi squared distance)&6.44\\
Wen et al.\cite{wen_model-based_2009}&5&RPF (HMM)&15.02\\
Vargas et al.\cite{vargas_off-line_2011}&5&LBP (SVM)&11.9\\
Vargas et al.\cite{vargas_off-line_2011}&10&LBP (SVM)&7.08\\
Ooi et al\cite{ooi_image-based_2016}&5&DRT + PCA (PNN)&13.86\\
Ooi et al\cite{ooi_image-based_2016}&10&DRT + PCA (PNN)&9.87\\
Soleimani et al.\cite{soleimani_deep_2016}&5&HOG (DMML)&13.44\\
Soleimani et al.\cite{soleimani_deep_2016}&10&HOG (DMML)&9.86\\
\hline

Proposed & 5 & SigNet (SVM) & 3.58 (+- 0.54) \\
\textbf{Proposed} & \textbf{10} & \textbf{SigNet (SVM)} & \textbf{2.87 (+- 0.42)} \\

Proposed & 5 & SigNet-F (SVM) & 3.70 (+- 0.79) \\
Proposed & 10 & SigNet-F (SVM) & 3.00 (+- 0.56) \\

\hline
\end{tabular}
}
\end{table}

\begin{table}
\centering
\caption{Comparison with the state-of-the-art in CEDAR}
\label{tbl:cedar_soa}
\resizebox{0.6\textwidth}{!}{%

\begin{tabular}{lrlc}
\hline
Reference & \# Samples & Features & AER/EER\\
\hline
Chen and Srihari\cite{chen_new_2006}&16&Graph Matching&7.9\\
Kumar et al.\cite{kumar_writer-independent_2010}&1&morphology (SVM)&11.81\\
Kumar et al.\cite{kumar_writer-independent_2012}&1&Surroundness (NN)&8.33\\
Bharathi and Shekar\cite{bharathi_off-line_2013}&12&Chain code (SVM)&7.84\\
Guerbai et al.\cite{guerbai_effective_2015}&4&Curvelet transform (OC-SVM)&8.7\\
Guerbai et al.\cite{guerbai_effective_2015}&8&Curvelet transform (OC-SVM)&7.83\\
Guerbai et al.\cite{guerbai_effective_2015}&12&Curvelet transform (OC-SVM)&5.6\\
\hline

Proposed & 4 & SigNet (SVM) & 5.87 (+- 0.73) \\
Proposed & 8 & SigNet (SVM) & 5.03 (+- 0.75) \\
Proposed & 12 & SigNet (SVM) & 4.76 (+- 0.36) \\

Proposed & 4 & SigNet-F (SVM) & 5.92 (+- 0.48) \\
Proposed & 8 & SigNet-F (SVM) & 4.77 (+- 0.76) \\
\textbf{Proposed} & \textbf{12} & \textbf{SigNet-F (SVM)} & \textbf{4.63 (+- 0.42)} \\

\hline
\end{tabular}
}
\end{table}

\begin{table*}
\centering
\caption{Comparison with the state-of-the-art on the Brazilian PUC-PR dataset (errors in \%)}
\label{table:soa_brazilian}
\ra{1.5}
\resizebox{\textwidth}{!}{%

\begin{tabular}{lllrrrrrrr}
\hline
Reference & \begin{tabular}[x]{@{}c@{}}\#samples\\per user\end{tabular}  & Features &  FRR &  FAR\textsubscript{random} &  FAR\textsubscript{simple} &  FAR\textsubscript{skilled} &  AER & AER\textsubscript{genuine + skilled} &  EER\textsubscript{genuine + skilled}  \\
\hline
Bertolini et al. \cite{bertolini_reducing_2010} & 15 &Graphometric & 10.16&3.16&2.8&6.48&5.65&8.32 & - \\
Batista et al. \cite{batista_dynamic_2012} & 30 & Pixel density & 7.5&0.33&0.5&13.5&5.46&10.5 & -\\
Rivard et al. \cite{rivard_multi-feature_2013} & 15 &ESC + DPDF &11&0&0.19&11.15&5.59&11.08& -\\
Eskander et al. \cite{eskander_hybrid_2013}& 30 &ESC + DPDF &7.83&0.02&0.17&13.5&5.38&10.67& -\\

\hline

  Present Work &                 5 &               SigNet &   4.63 (+- 0.55) &  0.00 (+- 0.00) &  0.35 (+- 0.20) &   7.17 (+- 0.51) &  3.04 (+- 0.17) &  5.90 (+- 0.32) &     2.92 (+- 0.44) \\
 Present Work &                15 &               SigNet &   1.22 (+- 0.63) &  0.02 (+- 0.05) &  0.43 (+- 0.09) &  10.70 (+- 0.39) &  3.09 (+- 0.20) &  5.96 (+- 0.40) &     2.07 (+- 0.63) \\
 \textbf{Present Work} &                \textbf{30} &               \textbf{SigNet} &   0.23 (+- 0.18) &  0.02 (+- 0.05) &  0.67 (+- 0.08) &  12.62 (+- 0.22) &  3.38 (+- 0.06) &  6.42 (+- 0.13) &     \textbf{2.01 (+- 0.43)} \\
 Present Work &                 5 &  SigNet-F  &  17.17 (+- 0.68) &  0.00 (+- 0.00) &  0.03 (+- 0.07) &   2.72 (+- 0.37) &  4.98 (+- 0.16) &  9.94 (+- 0.31) &     5.11 (+- 0.89) \\
 Present Work &                15 &  SigNet-F  &   9.25 (+- 0.88) &  0.00 (+- 0.00) &  0.25 (+- 0.09) &   6.55 (+- 0.37) &  4.01 (+- 0.24) &  7.90 (+- 0.46) &     4.03 (+- 0.59) \\
 Present Work &                30 &  SigNet-F  &   5.47 (+- 0.46) &  0.00 (+- 0.00) &  0.38 (+- 0.11) &   8.80 (+- 0.44) &  3.66 (+- 0.12) &  7.13 (+- 0.25) &     3.44 (+- 0.37) \\

\hline
\end{tabular}
}
\end{table*}

Tables \ref{tbl:mcyt_soa}, \ref{tbl:cedar_soa} and \ref{table:soa_brazilian} present the comparison with the state-of-the-art performance on MCYT, CEDAR and Brazilian PUC-PR, respectively. In all datasets we notice improvement in performance compared to the state-of-the-art, suggesting that the features learned on GPDS generalize well to signatures from other datasets (with different protocols for signature acquisition, created with different users in different countries). We also note that other methods proposed in the literature often present better performance only in one dataset, for instance, Guerbai et al. \cite{guerbai_effective_2015} obtained good results on CEDAR, but poor results on GPDS; Soleimani et al. \cite{soleimani_deep_2016} obtained good results on MCYT, but not on GPDS. The proposed method, however, obtained state-of-the-art performance in all datasets. For MCYT we obtained EER of 2.87\% compared to 6.44\% in the literature. On CEDAR, we obtained EER of 4.63\%, compared to 5.6\%. For the Brazilian PUC-PR dataset, we notice an improvement in performance both in terms of average error rate (considering all types of forgery), and the average error rate comparing only genuine signatures and skilled forgeries. It is worth noting that in these experiments we used a global threshold = 0 to report FRR and FAR, since we did not have a validation set to learn the appropriate global threshold, hence the large differences between FRR and FAR\textsubscript{skilled}.

We also noticed that the formulation that learned features using skilled forgeries from the GPDS dataset did not perform better in all cases. For MCYT and CEDAR the performance between SigNet and SigNet-F was not significantly different, whereas for the Brazilian PUC-PR dataset it obtained worse performance than SigNet. This suggests that the representation may have specialized to traits present in the forgeries made for the GPDS dataset, which depend on the acquisition protocol, such as if only one type of writing instrument was used, and the directions given to participants to create the forgeries. We note, however, that 1920 people participated in creating forgeries for the GPDS dataset \cite{vargas_off-line_2007}.

Finally, considering that the MCYT dataset contains both an Offline dataset (with static signature images, as used in this paper), and an Online version (with dynamic information of the strokes), it is possible to compare the two approaches to the problem. In the literature, online signature verification systems empirically demonstrate better performance than offline systems \cite{impedovo_automatic_2008}, which is often attributed to the lack of dynamic information of the signature writing process in the offline signatures. The gains in performance using the method proposed in this paper reduce the gap between the two approaches.  Using offline signatures, we obtained 2.87 \% EER\textsubscript{user thresholds} using 10 samples per user. Using online data, the best results reported in the literature achieve 2.85 \% EER \cite{rua_online_2012} and 3.36 \% EER \cite{fierrez_hmm-based_2007}, also using 10 samples per user. We note, however, that in our work we do not address the issue of selecting user-specific thresholds (or performing user-specific score normalization), which is left as future work. In constrast, both \cite{rua_online_2012} and \cite{fierrez_hmm-based_2007} use score normalization, followed by a single global threshold, so the comparison of these papers to our work is not direct.

\begin{figure}
\centering
\subfloat[GPDS-300]{
\includegraphics[width=0.45\textwidth]{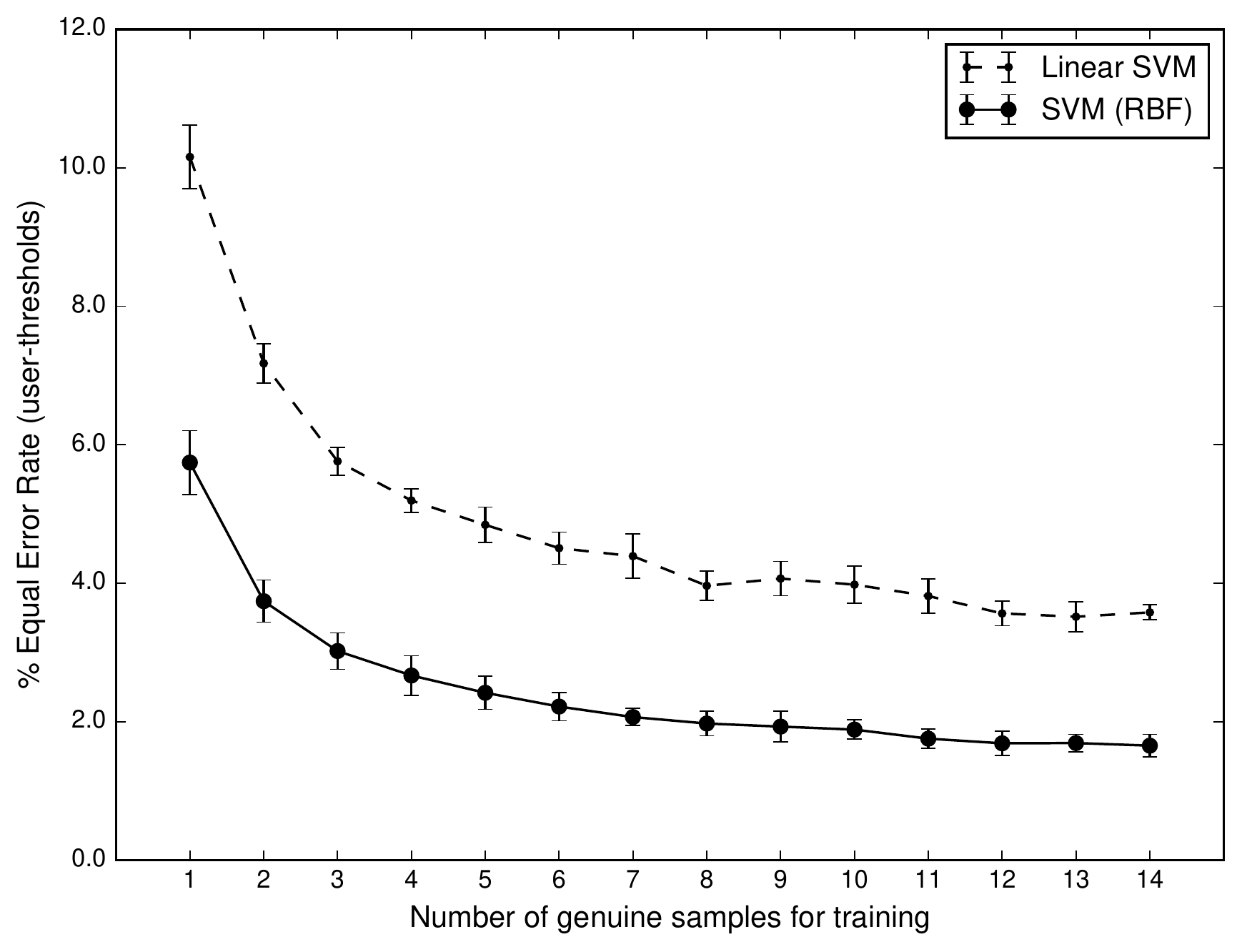}}
\qquad
\subfloat[MCYT]{
\includegraphics[width=0.45\textwidth]{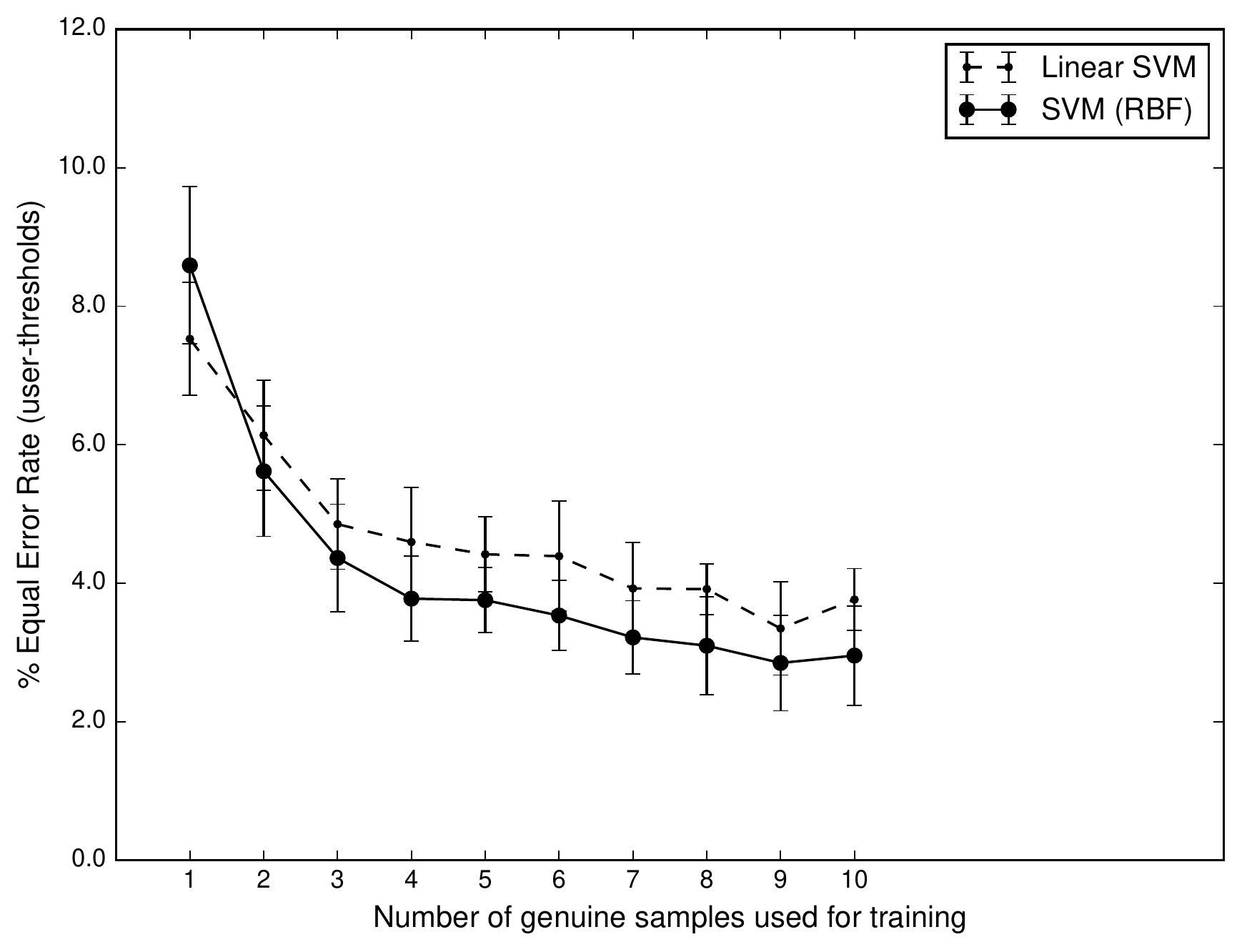}}
\qquad
\subfloat[CEDAR]{
\includegraphics[width=0.45\textwidth]{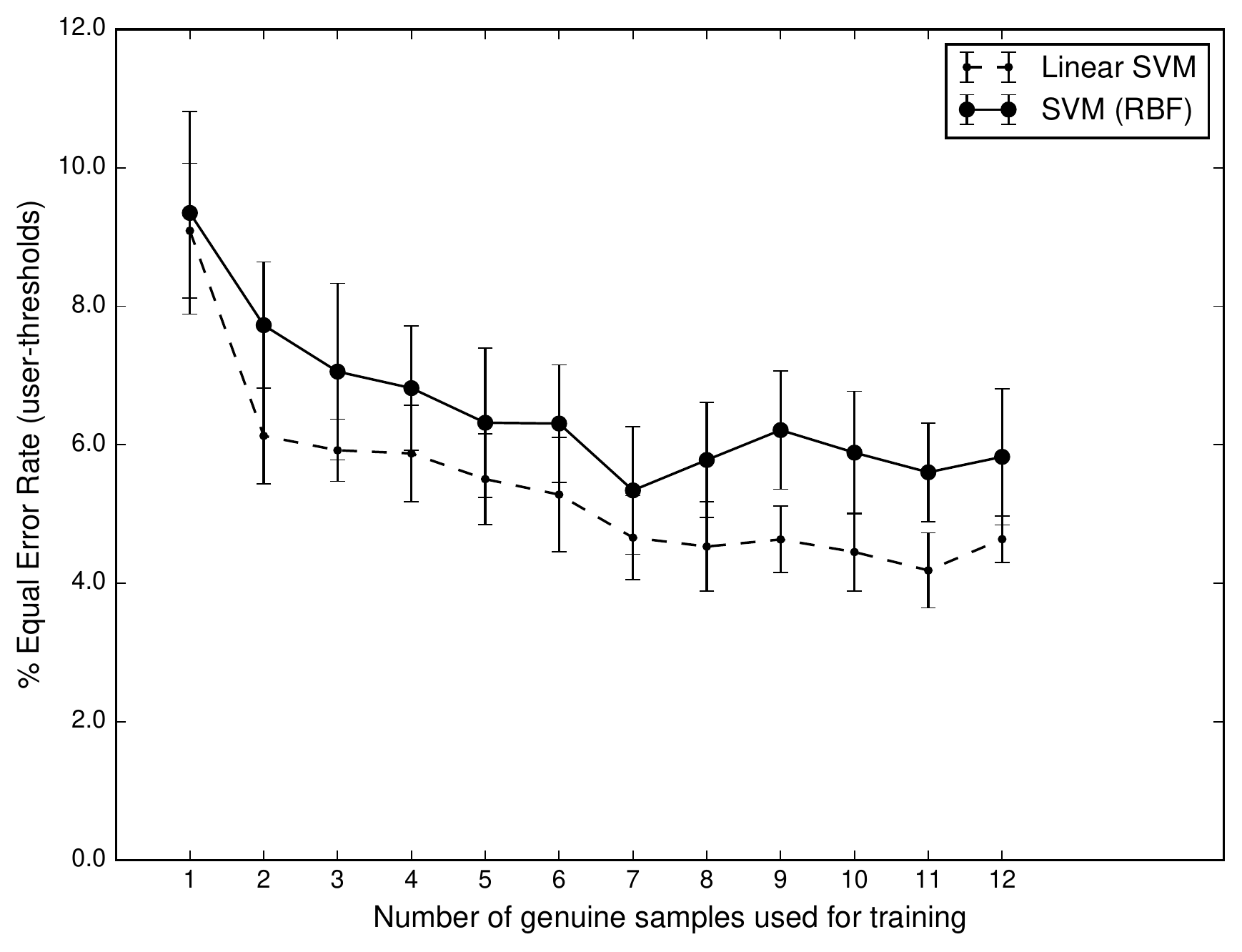}}
\qquad
\subfloat[Brazilian PUC-PR]{
\includegraphics[width=0.45\textwidth]{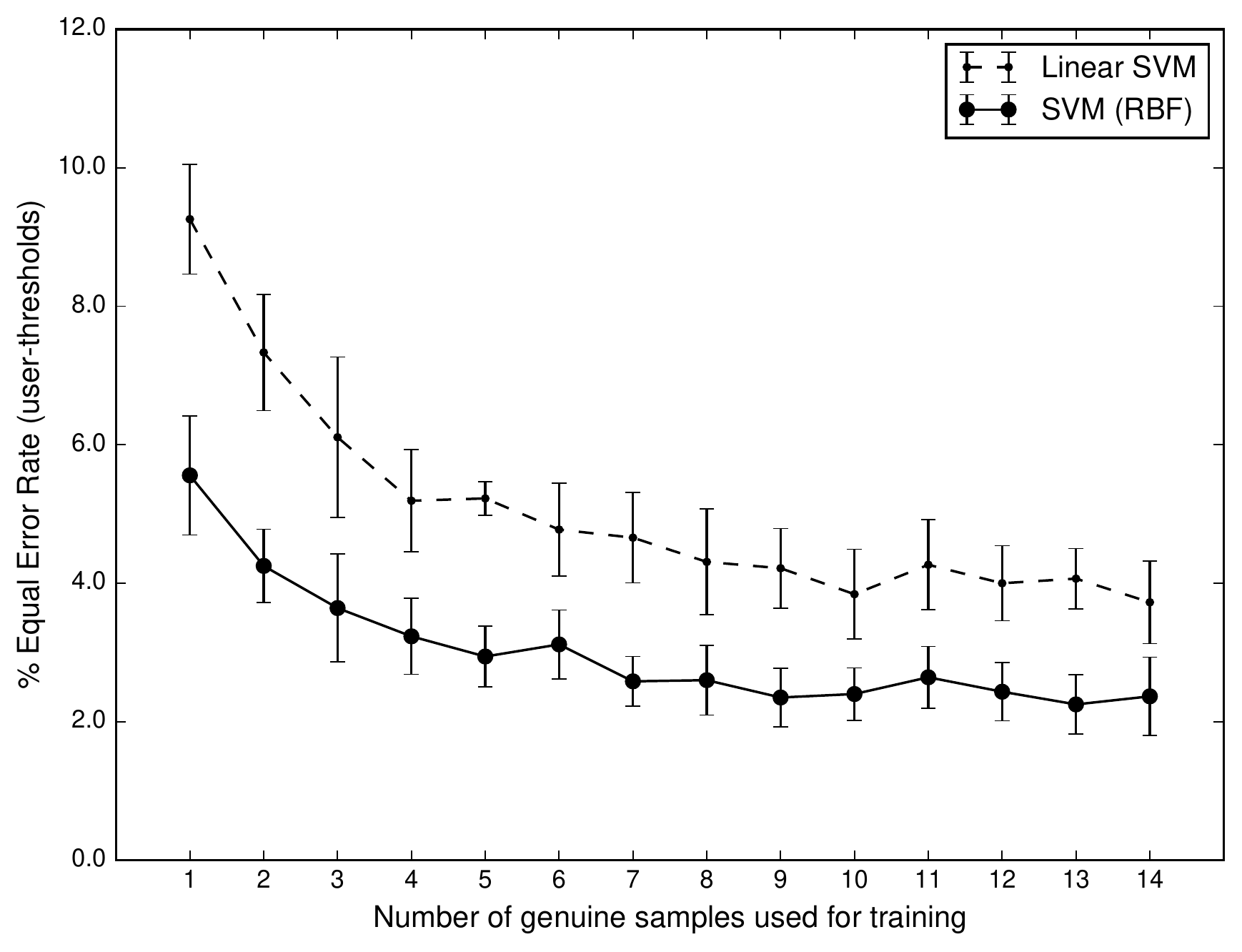}}
\qquad
\caption{Average performance of the Writer-Dependent classifiers for each dataset, as we vary the number of genuine signatures (per user) available for training.}
\label{fig:varying_users}
\end{figure}

\subsubsection{Varying the number of genuine samples available for training}

Figure \ref{fig:varying_users} shows the improvement in performance on the four datasets as we obtain more samples per user for training. Each point represents the performance of the WD classifiers trained with a given number of genuine samples (mean and standard deviation across 10 replications). As in previous work (\cite{eskander_hybrid_2013}, \cite{hafemann_ijcnn_2016}), we notice diminishing returns as we collect more samples for each user. It is worth noting that in the GPDS dataset, even with a single sample per user we obtain 5.74\% EER, which surpasses the state-of-the-art system that used 12 samples per user, showing that good feature representations are indeed critical to obtain good performance.

\section{Conclusion}
\label{sec:conclusion}

In this work, we presented different formulations for learning representations for offline signature verification. We showed that features learned in a writer-independent way can be very effective for signature verification, improving performance on the task, compared to the methods that rely on hand-engineered features.

In particular, we showed a formulation of the problem to take advantage of having forgery data from a subset of users, so that the learned features perform better in distinguishing forgeries for unseen users. With this formulation, we obtain an EER or 1.72\% in the GPDS-160 dataset, compared to 6.97\% reported in the literature. The visual analysis of the feature space shows that the features generalize well to unseen users, by separating genuine signatures and forgeries in different regions of the representation space. We also noted very good performance of this strategy even when few samples per user are available. For instance, with 5 samples per user, we obtained 2.41 \% EER on this dataset.

The experiments with the MCYT, CEDAR and Brazilian PUC-PR datasets demonstrate that the features learned in this Writer-Independent format not only generalize to different users of the GPDS dataset, but also to users from other datasets, surpassing the state-of-the-art performance on all three. We noticed, however, that the model learned with forgeries in the GPDS dataset did not perform better in all cases, suggesting that the characteristics of forgeries in the datasets may be different - this will be further studied in future work. Another promising research direction is the combination of online and offline signature verification methods. This can improve robustness of the system since it becomes harder to create a forgery that is misclassified by both classifiers, that is, a forgery having similar strokes in terms of speed of execution, and at the same time that is visually similar to a genuine signature from the user.

\section*{Acknowledgments}

This work was supported by the CNPq grant \#206318/2014-6 and by grant RGPIN-2015-04490 to Robert Sabourin from the NSERC of Canada.

\section*{References}

\bibliography{biblio}

\end{document}